# Selection of appropriate multispectral camera exposure settings and radiometric calibration methods for applications in phenotyping and precision agriculture

**Vaishali Swaminathan[a,*], J Alex Thomasson[b,a], Robert G Hardin[a], Nithya Rajan[c]**

*Corresponding author: Vaishali Swaminathan (vaishaliswaminathan@tamu.edu)

[a] Department of Biological and Agricultural Engineering, Texas A&M University, College Station, Texas, USA

[b] Department of Agricultural and Biological Engineering, Mississippi State University, Starkville, Mississippi, USA

[c] Depart of Soil and Crop Sciences, Texas A&M University, College Station, Texas, USA

## Abstract

Radiometric accuracy of data is crucial in quantitative precision agriculture, to produce reliable and repeatable data for modeling and decision making. The effect of exposure time and gain settings on the radiometric accuracy of multispectral images was not explored enough. The goal of this study was to determine if having a fixed exposure (FE) time during image acquisition improved radiometric accuracy of images, compared to the default auto-exposure (AE) settings. This involved quantifying the errors from auto-exposure and determining ideal exposure values within which radiometric mean absolute percentage error (MAPE) were minimal (< 5%). The results showed that FE orthomosaic was closer to ground-truth (higher $R^2$ and lower MAPE) than AE orthomosaic. An ideal exposure range was determined for capturing canopy and soil objects, without loss of information from under-exposure or saturation from over-exposure. A simulation of errors from AE showed that MAPE < 5% for the blue, green, red, and NIR bands and < 7% for the red edge band for exposure settings within the determined ideal ranges and increased exponentially beyond the ideal exposure upper limit. Further, prediction of total plant nitrogen uptake (g/plant) using vegetation indices (VIs) from two different growing seasons were closer to the ground truth (mostly, $R^2 > 0.40$, and MAPE = 12 to 14%, $p < 0.05$) when FE was used, compared to the prediction from AE images (mostly, $R^2 < 0.13$, MAPE = 15 to 18%, $p >= 0.05$).

# Highlights

1. Determined the use of default auto-exposure settings on multispectral cameras affected radiometric accuracy.
2. Fixed exposure setting for a Micasense RedEdge-3 camera was determined.
3. The object-based ELM calibration worked on the auto-exposure and fixed-exposure images where the focus was on plant and soil reflectance.
4. Fixed exposure multispectral camera settings resulted in better spatiotemporal radiometric accuracy than auto-exposure settings.

## 1. Introduction

Multispectral cameras are widely used in agricultural remote sensing due to their ability to capture a wide range of information – spectral signatures, spatial patterns, structural details, temporal variations – especially when mounted on platforms that provide fast-coverage like unmanned aerial vehicles (UAVs), piloted aircraft, and satellites. They often operate at relatively narrow (e.g., 10 to 40 nm) bandwidths that are sensitive to plant-canopy and soil reflectance and usually include the visible (blue, green, red) and infrared (red edge, NIR, SWIR, etc.) regions of the spectrum. Some popular multispectral cameras for agriculture are the Micasense RedEdge and Altum series, the Parrot Sequoia, Sentera sensors, and the DJI P4, all of which have been used extensively on UAVs. Since UAV-based multispectral images are captured from higher altitudes, they are subjected to atmospheric attenuation and reduction in spatial resolution, for which calibration is necessary.

Multispectral data can be classified as qualitative or quantitative based on the application. Examples of qualitative applications are spatial pattern recognition for disease detection [1] and detection and counting of fruiting organs, plants, and invasive species [2-4], where contrast, brightness, sharpness, and resolution are more important than the exact values of the pixels. Quantitative applications involve extracting measurable plant spectral and morphological features for precision input management [5-7], yield estimation [2,8,9], and phenotyping [10,11]. Quantitative spectral features include spectral/radiometric features like band reflectance measures and vegetation indices (VIs). The light reflected from field crops needs to be estimated accurately to perform reliable quantitative analysis, but it tends to change with season, weather, and time of the day.

Radiometric calibration is the standardization of images acquired from various sensors used at different locations and times under various environmental conditions. The methods adopted to minimize effects of spatiotemporal and environmental factors on UAV multispectral images include scheduling data collection under clear sky or constant illumination within a timeframe close to solar noon to minimize long shadows and intercept irradiance and reflectance perpendicular to the camera irrespective of the location. Raw digital numbers (DNs) that are camera-specific (dependent on light sensitivity or ISO, radiometric resolution, focal length, etc.) are converted to radiance and reflectance. Radiance is a measure of light reflected by objects in the scene, and reflectance is a unitless ratio of object-reflected radiance to solar irradiance. Hence, accurate determination of atmospheric attenuation factors, lens intrinsics(focal length and principal point), lens distortion, and other camera related parameters (pixel size, resolution, vignette model, exposure settings), and is necessary for accurate and precise determination of radiance or reflectance. periodic in-lab lens calibration using calibration frames [12] and integrating spheres [13,14] were recommended to account for changes in camera intrinsic parameters and image vignette and row gradient, respectively, due to mechanical wear and tear from repeated usage and harsh conditions. Cao et al. (2020) [15] used a look-up table approach to convert DNs from a RedEdge camera to radiance.

Radiometric calibration with ground truth values obtained from known-reflectance targets with approximately Lambertian reflectance is widely used to correct atmospheric effects in UAV images and other errors not accounted for by camera calibration. This method involves deriving a linear relationship between known and estimated radiance or reflectance of the targets and is broadly termed as the empirical line method (ELM) [16] or linear regression method (LRM) [17], and it varies depending on the number of targets used. Mamaghani and Salvaggio (2019a) [13] described 1-point and multi-point ELM based on manufacturer-provided calibration panels and in-field color gradient calibration targets, respectively. Camera manufacturers generally provide a single calibration panel to calibrate reflectance based on images of panels captured at ground level before and after UAV flights. However, this method does not account for atmospheric attenuation, camera dark current, and lens imperfections. A comparative study of different radiometric calibration methods on Parrot Sequoia images found that the multi-point ELM methods applied in post processing calibration of orthomosaics were the most accurate [18]. Guo et al. (2019) [17] noted that atmospheric attenuation of reflectance increased with flight altitude, and three color-graded (dark, moderate, white) reflectance targets were sufficient for the MCA camera used in their study. Chakhvashvili et al. (2021) [19] used nine panels with gradients ranging from black to white to calibrate a Micasense dual camera system and concluded that multi-point ELM produced 50% lower errors than 1-point ELM in estimating reflectance. Sub-band ELM calibration has been described as a band-wise multi-point ELM method in which the low-reflecting visible bands are calibrated with a power model and the high-reflecting infrared bands are calibrated with linear models [20]. Sub-band ELM was generalized by Luo et al. (2022) [21] and named piecewise-ELM, in which power and linear models for all bands were chosen depending on the reflectance range of the objects in the image. These studies did not probe whether the non-linear relationship between image pixel values and the reflectance of calibration targets in a given band were due to saturation from over-exposure, which can happen if the objects in the scene were predominantly of low reflectivity.

The spectral reflectance of field crops during the vegetative and reproductive phases were found to be < 25 % and 25 to 90 % in the visible and infrared bands, respectively [22-24]. Likewise, soil reflectance is generally low (< 25 %) in the visible bands and varies somewhat with changes in moisture content, but reflectance in the IR region is greatly influenced by moisture content, with dry soil reflecting > 25 % and moist soil reflecting < 15 % [25]. Zhang et al. (2012) [23] also noted that spectral characteristics of crops can dynamically change over time, wherein a 40 % increase in green reflectance occurred in the post-reproductive growth stages. Therefore, in agricultural fields, where plants and soil predominantly constitute the scene, bright pixels like those of white calibration targets tend to be saturated (information loss), especially in the visible bands. Olsson et al. (2021) [26] observed pixel saturation of calibration panels in their study and recommend installing larger panels that can prompt cameras to adjust their exposure settings and thus the dynamic range to accommodate the high-reflectance objects. However, this may result in datasets with very large variations in exposure time and gain.

Among the various camera related parameters stated previously, the effect of exposure settings on radiometric accuracy of images is the least researched subject. The default and widely used automatic exposure (auto-exposure) settings in cameras are programmed to optimize the dynamic range of the overall scene captured by them. Bagnall et al. (2023) [27] demonstrated that UAV-based RedEdge-3 images with fixed gain and exposure-time settings had lower radiometric calibration errors compared to images obtained with auto-exposure settings, but the reason behind the differences needs elucidation. It is currently unclear how variations in exposure time and gain affect radiometric calibration accuracy.

The overall goal of this study was to further clarify whether fixed camera exposure settings provide more consistent and accurate estimates of reflectance from UAV-borne multispectral cameras compared to the usually implemented auto-exposure settings, as well as to posit reasons for any differences found. The goal was broken down into four objectives: (1) to determine the ideal camera settings to perform ELM for each band by gradually varying the camera's exposure time and gain, (2) to check for potential errors from autoexposure by cross-calibration (i.e., applying ELMs derived from different exposure time-gain combinations on calibration target images), (3) to compare spatiotemporal radiometric accuracies of the fixed and auto-exposure orthomosaics based on distributed in-field reflectance calibration targets, and (4) to assess the practical implications of accuracy differences between fixed and auto-exposure settings by comparing total nitrogen (N) estimated from multiple VIs obtained from fixed and auto-exposure images.

## 2. Methods

### 2.1 Experiment Setup

A RedEdge-3 five-band multispectral camera with a downwelling light sensor (DLS-2) (AgEagle Aerial Systems, Wichita, KS, USA) was mounted on a Matrice 100 UAV (DJI, Shenzhen, Guangdong, China) and used to collect aerial images from a cotton field at the Texas A&M Agrilife Research farm near College Station, TX, USA (Latitude: 30.550338, Longitude: - 96.435214). The camera consisted of five imaging detectors with narrow-band optical filters having central wavelength and full-width at half-maximum (FWHM) range as follows: blue (475 nm; 20 nm), green (560 nm; 20 nm), red (668 nm; 10 nm), red edge (RE: 717 nm; 10 nm), and near infrared (NIR: 840 nm; 40 nm). The DLS-2 (referred to as DLS hereafter) included a light sensor and GPS/IMU unit and was synchronized with the camera to record irradiance at each waveband along with geographic coordinates at the time of image capture. A calibration reflectance panel (CRP) (Figure 1 right) was provided by the camera manufacturer for 1-point linear calibration of the multispectral images.

Multiple in-field reflectance calibration targets (60 cm x 60 cm) (Figure 1 left) were distributed across the field to verify the accuracy of reflectance estimated from aerial multispectral images. Each ground reference target was composed of black (B, low reflective), gray (G, medium reflective), and white (W, high reflective) rubber mats coated with matte-finish paints to minimize non-Lambertian (specular) reflection. The reflectance values of the calibration targets were measured a few days prior to the flights with a hand-held ASD spectroradiometer (FieldSpec-4, Analytical Spectral Device Inc., Boulder, CO, USA). The reflectance values collected with 1 nm resolution from 350 nm to 2500 nm from the spectroradiometer were averaged over the bandwidth corresponding to the FWHM of the bands (Table 1). Additionally, ground control points (GCPs) were placed inside the field for georectification, and their coordinates were measured with < 2 cm accuracy based on RTK GNSS (Reach RS2, Emlid, Budapest, Hungary) measurements.

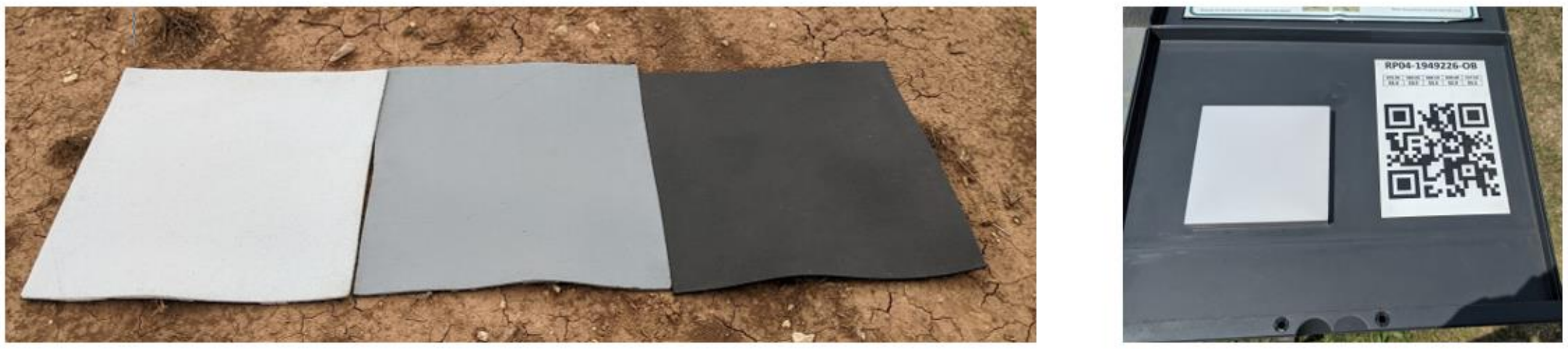

**Figure 1 In-field reflectance calibration targets with white (W, high), gray (G, medium), and black (B, low) reflectance calibration targets (left) and Micasense calibrated reflectance panel (CRP) (right).**

**Table 1 ASD FieldSpec-4 spectroradiometer measured average reflectance ratios of the black (B), gray (G), and white (W) in-field calibration targets in the blue, green, red, red edge, and NIR bands for years 2021 and 2022.**

| Bands | Year 2021 | | | Year 2022 | | |
|---|---|---|---|---|---|---|
| | B | G | W | B | G | W |
| Blue | 0.08 | 0.28 | 0.53 | 0.08 | 0.33 | 0.86 |
| Green | 0.08 | 0.26 | 0.50 | 0.08 | 0.31 | 0.86 |
| Red | 0.08 | 0.23 | 0.46 | 0.08 | 0.28 | 0.84 |
| Red edge | 0.09 | 0.23 | 0.45 | 0.08 | 0.27 | 0.82 |
| NIR | 0.09 | 0.21 | 0.42 | 0.08 | 0.25 | 0.85 |

## 2.2 Data Collection

### 2.2.1 Exposure time and gain sweep from a stationary UAV

To study the effect of camera brightness parameters (exposure time and gain) on reflectance estimation, an exposure sweep experiment was conducted in which the UAV was controlled to hover at 30 m above ground level (AGL) over one set of in-field reflectance calibration targets under clear sky conditions within ± 30 mins of local solar noon (13:30 hours, approx.), and the camera exposure was remotely adjusted across 15 values from 0.067 ms to 4.30 ms and 0.0067 ms to 2.00 ms, for gain values 1x and 2x, respectively. Exposure and gain for the five bands were adjusted independently. Just before the UAV was deployed, images of the CRP were also collected close to ground level by holding the UAV over the panel while changing the exposure and gain settings manually to the same values that were used for UAV image acquisition.

### 2.2.2 UAV flights

The UAV flight missions were executed with UgCS mission planning software (SPH Engineering, Riga, Latvia) at 30 m AGL with > 80 % forward and > 70 % lateral overlaps. The ground sampling distance of the camera at 30 m AGL was 2.08 cm. The flights were conducted within ±1 hour of solar noon to minimize shadows and bidirectional reflectance and under clear conditions to maintain consistent illumination. One flight was conducted with the camera's auto-exposure settings, and another was conducted with manually fixed exposure settings, determined for each band from the exposure sweep experiment described previously. Fixed exposure datasets were acquired with 1x gain, and the exposure settings for the five bands in each dataset are listed in Table 2.The fixed and auto-exposure images were collected on 17 June 2021 and 16 June 2022 from a cotton field when the crops were at similar phenological stages. The 2021 dataset had no variations in illumination, while the 2022 auto-exposure dataset had minor changes in illumination during the flight due to a thin layer of moving cloud. Further, the agronomic relevance of the auto-exposure and fixed exposure image collection methods were considered. Ground truth data from a nitrogen management trial within the cotton field were referenced, and destructive biomass analysis through combustion was performed to derive total biomass nitrogen (N) content, which was then used as a parameter to assess the performance of fixed and autoexposure images in precision N management.

**Table 2 Exposure settings applied to the blue, green, red, red edge, and NIR bands on the 17 June 2021 and 16 June 2022 fixed exposure flights.**

| Wavelength Bands | Exposure settings (ms), gain 1x | |
|---|---|---|
| | 17 June 2021 | 16 June 2022 |
| Blue | 0.58 | 0.76 |
| Green | 0.58 | 0.58 |
| Red | 0.76 | 0.76 |
| Red edge | 1.00 | 0.76 |
| NIR | 1.00 | 0.76 |

## 2.3 Data processing and analysis

### 2.3.1 Reflectance from raw images

The individual images of the in-field reflectance target obtained from the UAV were processed with Python based on the Micasense image processing library (GitHub repo: https://github.com/micasense/imageprocessing/tree/master/micasense). The raw image data (digital numbers, or DN) from the camera were converted to the quantitative measure of radiance (L, W/m2/sr/nm), denoting the amount of radiation reaching the camera from ground level. The conversion was based on the camera's intrinsic parameters, brightness settings, a pre-determined vignette model, and calibration coefficients stored in the image metadata (equations 1 to 4).

$$L = V(x,y) * \frac{a_1}{g} * \frac{P - \mathrm{P_{BL}}}{t_e + a_2 * y - a_3 * t_e * y} \quad (1)$$

$$V(x,y) = \frac{1}{k} \quad (2)$$

$$k = 1 + k_0 r + k_1 r^2 + k_2 r^3 + k_3 r^4 + k_4 r^5 + k_5 r^6 \quad (3)$$

$$r = \sqrt{(x - C_x)^2 - \left(y - C_y\right)^2} \quad (4)$$

where,

V(x,y) was the vignette model

x, y represented the pixel coordinates

P and $P_{BL}$ was the normalized raw pixel and black level values

$g$ and $t_e$ were gain and exposure time

$a_1$, $a_2$, $a_3$, were radiometric calibration coefficients

$k_0$, $k_1$, $k_2$, k3, $k_4$, $k_5$ were camera's vignette correction coefficients

r was the Euclidean distance of pixel (x,y) from the center

$C_x$, $C_y$ were the center pixel coordinates

The CRP based calibration was performed to offset intrinsic camera errors. This is an optional step, as intrinsic errors can be compensated along with atmospheric errors in post-processing calibration. The radiance values of the CRP images in all bands were converted to nominal reflectance ($\rho_{CRP}$) based on the DLS recorded direct irradiance ($E_{CRP}$, µW/cm2/nm) at the time of image acquisition (equation 5). Then a correction factor (F) for each band was obtained by dividing estimated CRP reflectance by known reflectance ($\rho_{CRP_known}$), providing 1-point ELM calibration equations of all images in the dataset (equation 6).

$$\rho_{CRP} = 100 * \pi * \frac{L_{CRP}}{E_{CRP}} \tag{5}$$

$$F = \frac{\rho_{CRP}}{\rho_{CRP_known}} \tag{6}$$

The correction factor was applied to the UAV images to obtain a CRP-calibrated reflectance value (equation 7).

$$\rho = 100 * \pi * F * \frac{L}{E} \tag{7}$$

Here E is the at-altitude irradiance measured by the DLS light sensor when the UAV images were captured. Note that it was important to convert radiance to reflectance for further analysis, because the ground-truth reference data obtained from the spectroradiometer were also in terms of reflectance. The Python workflow implemented parallel processing through multi-threading to improve the processing speed of the script.

An OpenCV script was deployed to draw square bounding boxes that extracted average estimated reflectance of the black, gray, and white targets upon manual identification of target centers. The estimated reflectance of the targets was compared to spectrometer-measured reflectance to analyze the impact of exposure and gain settings on reflectance estimation accuracy and dynamic range of the images, which indicated the radiometric resolution captured by the camera. This information was used to derive ideal exposure time and gain settings to perform full scale multipoint ELM that covered the entire reflectance scale from 0 to 1 and object-based ELM calibration, where the calibration range was customized to focus on the reflectance of canopy, soil, and other objects of interest (described below under post-processing calibration).

A set of five images was sampled for each exposure-gain setting combination and categorized into reference and target sets. While both sets had the same images, the reference set was used to derive object-based ELM, and the target set was used to assess the errors from ELM calibration. We use the term 'cross-calibration' to describe this process, which was performed to quantify radiometric calibration errors from variations in exposure and gain during auto-exposure flights. Mean absolute percentage error (MAPE) < 5 % was used as the acceptable threshold to test whether object-based ELM achieved better results than similar studies performed by Deng et al. (2018) [20] and Luo et al. (2022) [21], who had errors over 10 %.

#### 2.3.2 Photogrammetric processing and post-processing calibration

Photogrammetric processing of the auto-exposure and fixed-exposure UAV flight images was performed in Metashape Pro software (Agisoft LLC, St. Petersburg, Russia). The first step in the workflow was tie-point matching, which used tie-point matching algorithms like SIFT, SURF, ORB, etc., in conjunction with structure from motion (SfM), the details of which are not disclosed by the software developers. The GCPs were manually identified in the aerial images and tagged for geo-rectification of the scene. The geo-rectified tie-points were filtered to remove outliers, and a densified 3-D point-cloud was obtained and used to generate a high-resolution digital elevation model (DEM) and a 5-band orthomosaic of the entire field. The highest possible accuracy settings in Metashape Pro were used for each step in the workflow. An initial 1-point ELM with the CRP panel images was performed to convert the raw DNs to roughly estimated reflectance before generating orthomosaic images. As stated in the previous section, this conversion was performed to remove camera induced errors and was an optional step, as camera errors could have been accounted for in post-processing calibration, discussed below.

The orthomosaic images were calibrated in post-processing to remove atmospheric noise in reflectance estimation and camera biases if CRP-based 1-point ELM was not performed. The in-field reflectance calibration targets were manually identified, and average reflectance of each reflectance target was obtained from shapefile polygons (35 cm x 35 cm) situated at the center of each target. An object-based ELM approach (Table 3) was used to calibrate the individual bands of the orthomosaic in postprocessing. The object-based ELM was adopted from the sub-band ELM method of Deng et al. (2018) [20] and the piecewise ELM approach of Luo et al. (2022) [21], in which calibration equations were determined to be linear or power models based on whether the objects studied were high or low reflecting. In the previous studies, both methods used auto-exposure settings for image acquisition, performed calibration on the entire dynamic range (black to white) for all bands, and applied non-linear calibration equations in low-reflecting bands. The object-based ELM implemented in the current study was linear, and the calibration was performed by selecting targets that tightly bounded the expected range of soil and canopy reflectance (Table 3). For example, the green band was calibrated between black and gray targets due to the low green reflectance of canopy and soil, while the NIR band was calibrated with black, gray, and white targets due to higher NIR reflectance of both soil and canopy.

**Table 3 Object-based calibration range for the five bands of the Micasense RedEdge-3 multispectral camera set using the black (B), gray (G), and white (W) color gradients of the in-field reflectance calibration targets.**

| Wavelength Bands | Object-based calibration range |
|---|---|
| Blue | B – G |
| Green | B – G |
| Red | B – G |
| Red edge | B – G – W |
| NIR | G – W |

Images from the auto-exposure flight dataset were selected at random and segregated into three categories: predominantly canopy, predominantly soil, and in-field reflectance calibration targets. The exposure time and gain distribution in each category was studied. To verify the spatiotemporal uniformity of reflectance estimated from the auto and fixed exposure flights, averaged reflectance values of the calibration targets obtained from the orthomosaics were compared to spectroradiometer measured reflectance values. Spatial uniformity (precision) of reflectance was determined by the coefficient of determination ($R^2$), which indicated the goodness of fit of the target reflectance along the empirical line obtained from regression The mean absolute percentage error (MAPE) after post-processing calibration was used to determine spatiotemporal accuracy in reflectance estimation as compared to spectroradiometer measured values.

$$R^2 = 1 - \frac{\sum_{i=1}^{n}(\hat{y}_i - y_i)^2}{\sum_{i=1}^{n}(y_i - \bar{y}_i)^2} \tag{8}$$

$$MAPE = \frac{1}{n}\sum_{i=1}^{n}\left|\frac{\hat{y}_i - y_i}{y_i}\right| \times 100 \tag{9}$$

### 2.3.3 Vegetation Index for Total N Estimation

To test the performance of fixed and autoexposure settings in real world applications – specifically, in a cotton nitrogen management study – VIs (Table 4) were computed from the respective orthomosaics. The seven VIs are sensitive to canopy structure, leaf chlorophyll, and nitrogen content, all of which are indicators of accumulated total nitrogen in the plants (g/plant). The VIs from fixed and autoexposure data obtained on 17 June 2021 and 16 June 2022 were compared with total biomass N. These dates represented similar growth stages with data acquired under different conditions to validate the consistency of the calibration methods. Linear regression was performed between plot-level averages of the VIs and plot-level total N . The $R^2$ and MAPE were used as metrics to assess the performance of the two exposure-setting methods in estimating total biomass N content in cotton.

**Table 4 Vegetation indices (VI) and their formula for monitoring cotton nitrogen status based on observed reflectance (ρ) in the blue, green, red, red edge, and NIR bands of the RedEdge-3 multispectral camera.**

| Index | Formula |
|---|---|
| Normalized difference vegetation index (NDVI) | $\frac{\rho_{NIR} - \rho_{red}}{\rho_{NIR} + \rho_{red}}$ |
| Normalized difference red edge index (NDRE) | $\frac{\rho_{NIR} - \rho_{rededge}}{\rho_{NIR} + \rho_{rededge}}$ |
| Triangular greenness index (TGI) | $-0.5\left[(668 - 475)(\rho_{red} - \rho_{green}) - (668 - 560)(\rho_{red} - \rho_{blue})\right]$ |
| Green Normalized difference Vegetation Index (GNDVI) | $\frac{\rho_{NIR} - \rho_{green}}{\rho_{NIR} + \rho_{green}}$ |
| Chlorophyll index red edge ($CI_{rededge}$) | $\frac{\rho_{NIR}}{\rho_{rededge}} - 1$ |
| Chlorophyll Index green ($CI_{green}$) | $\frac{\rho_{NIR}}{\rho_{green}} - 1$ |
| Renormalized difference vegetation index (RDVI) | $\frac{\rho_{NIR} - \rho_{red}}{\sqrt{\rho_{NIR} + \rho_{red}}}$ |

## 3. Results

### 3.1 Exposure and gain variations in auto-exposure flights

Large variations in exposure time and gain were observed when auto-exposure settings were used for capturing different field objects (in-field reflectance calibration targets, predominantly canopy, and predominantly soil pixels) at 30 m AGL on 17 June 2021 (Figure 2) and 16 June 2022 (Figure 3), respectively. The red edge band had the least variability in exposure time and gain, followed by the NIR and green bands. The most variability in exposure and gain were in the blue and red bands. Overall, there was more variability in exposure and gain in the 2022 dataset than the 2021 dataset for all bands. The red edge band exposure settings were similar (< ± 0.10 ms) among the three objects for both years, although the 2022 dataset had marginally higher variability. The NIR band exposure distribution in 2021 was less than 1.00 ms, but in 2022 the values were predominantly greater than 1.00 ms for all categories. The green exposure values for both years were similar, although the variations were greater in 2022. The gain value for the red edge band was predominantly 2x, and the green and NIR gains were mostly 1x. In 2021 , the red band exposure variability was minimal and differences in exposure were observed only for soil category, but in 2022 more pronounced variations in exposure time and gain were observed. The blue band displayed the most drastic changes in exposure and gain among all three object categories, and these changes were more pronounced in the 2022 images. For 2021 and 2022 combined, the exposure time ranges for the blue band were 1.00 to 2.00 ms (1x gain) and 1.00 to 1.30 ms (2x gain). The combined exposure ranges for the green, red, red edge, and NIR bands were 0.80 to 1.40 (1x), 1.40 to 2.00 (1x), 1.00 to 1.20 (2x), and 0.90 to 1.30 ms (1x), respectively.

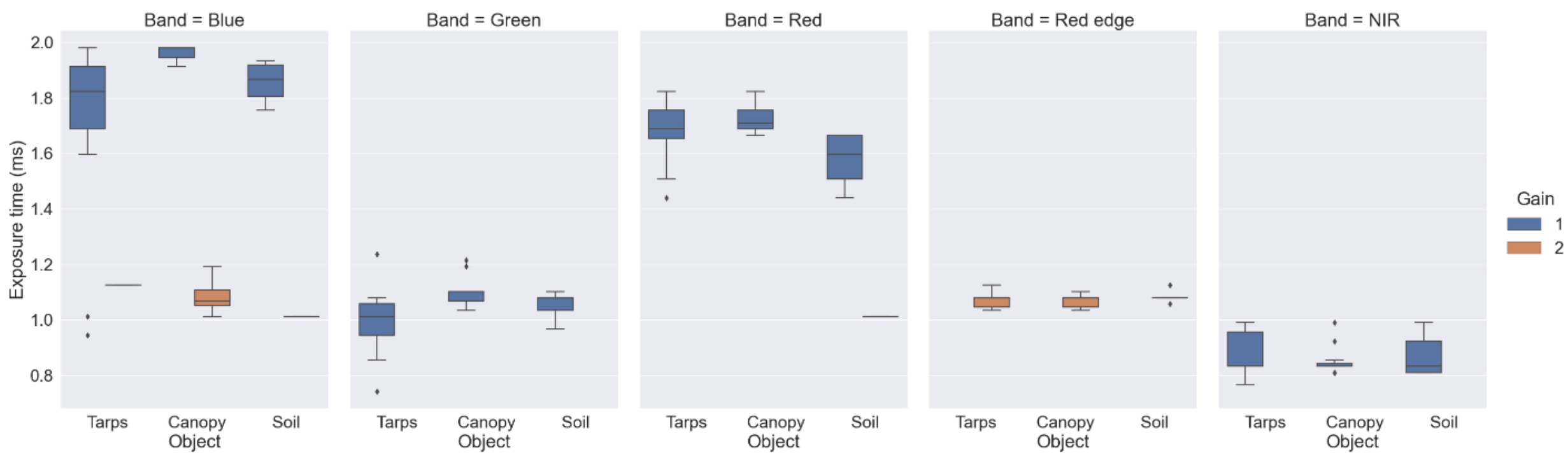

**Figure 2 Exposure and gain distribution for images containing in-field reflectance calibration targets (tarps), predominantly canopy, and predominantly soil pixels in each of the five bands of the RedEdge-3 multispectral camera, collected on 17 June 2021 during a 30 m AGL autoexposure UAV flight. Only the blue band was automatically captured with 1x and 2x gains for all objects. All other bands were captured either completely with 1x (green, NIR) or 2x (red edge) or with only some objects having both 1x and 2x (red) gains.**

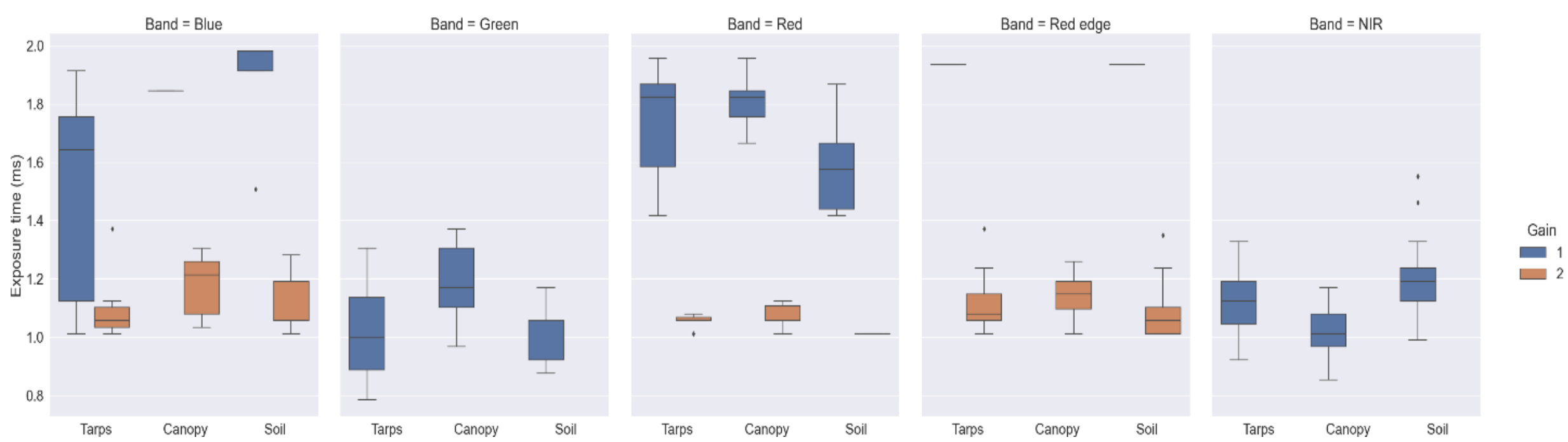

**Figure 3 Exposure and gain distribution for images containing in-field reflectance calibration targets (tarps), predominantly canopy, and predominantly soil pixels in each of the five bands of the RedEdge-3 multispectral camera, collected on 16 June 2022 during a 30 m AGL autoexposure UAV flight. Only the blue and red bands were automatically captured with 1x and 2x gains for all objects. All other bands were captured either completely with 1x (green, NIR) or with only some objects having both 1x and 2x (red edge) gains.**

### 3.2 Reflectance response to exposure time and gain

The graphs in Figure 4 and Figure 5 depict the variations in estimated reflectance of the black, gray, and white calibration targets in response to the exposure sweep experiment for gains 1x and 2x, respectively. The estimated white target reflectance across all bands was observed to be marginally lower than the spectroradiometer measured white target reflectance, possibly due to dust accumulation on the target surfaces under field conditions. It is also likely that the CRP calibrated UAV images were subject to atmospheric distortion that resulted in lowered estimates of the white targets. It was observed that increasing the exposure time beyond a threshold (blue and red vertical dotted lines in Figure 4 and Figure 5 ) decreased the dynamic range/resolution of the estimated reflectance and gradually resulted in saturation at 0.50 reflectance ratio. It was also noted that divergence and saturation occurred faster for the highly reflective white targets, followed by the gray and black targets. For example, the divergence in estimated reflectance from ground truth in the blue band was first observed for the white target with 1x gain at approximately 0.50 ms, then the gray target at 1.00 ms, and finally the black target at 1.40 ms. Also, saturation for all three reflectance targets occurred at lower exposure times as the gain increased from 1x to 2x. Table 5 lists ideal exposure settings under 1x and 2x gains for the Micasense RedEdge-3 multispectral camera at 30 m AGL in order to perform full-scale (red dotted lines in Figure 4 and Figure 5) and object-based (blue dotted lines in Figure 4 and Figure 5) reflectance calibration. For object-based image acquisition, the upper limits of exposure time for the visible bands were selected such that the estimated gray target reflectance did not deviate from the ground truth and the white target reflectance did not saturate at 0.50. The upper limits of object-based captures are generally higher than full-scale captures, where the limits are set by the white target. The upper limits of the red edge and NIR bands were the same for full-scale and object-based ELM due to the wider range of canopy and soil reflectance observed in them. The thresholds for red edge and NIR were chosen such that the estimated reflectance of all the targets remained close to ground-truth. While the estimated reflectance of all three targets remained consistent at extremely low exposure times, lower limits were set to avoid loss of details due to under-exposure of low reflecting canopy and soil objects.

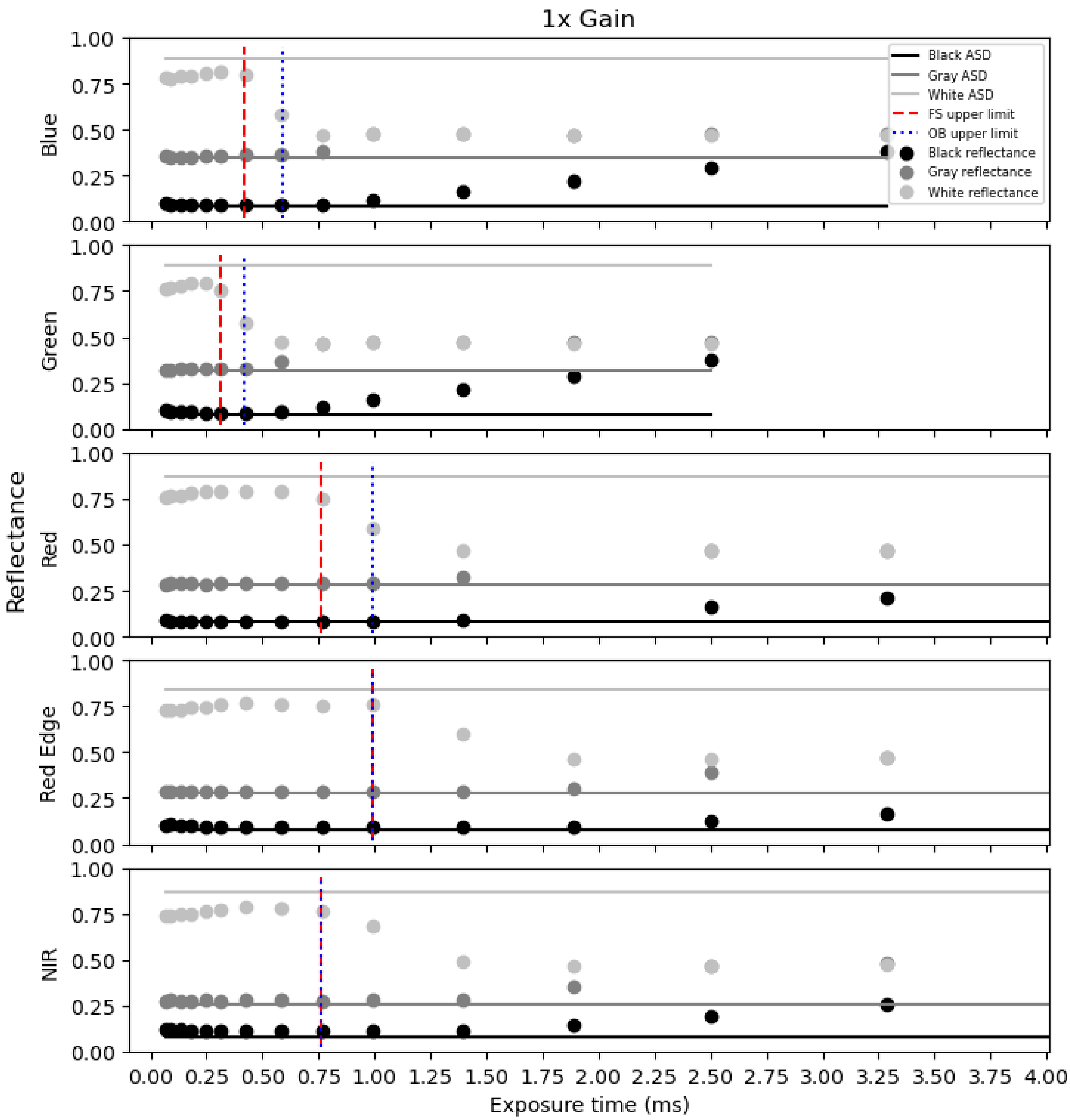

**Figure 4 Reflectance estimation of the in-field reflectance calibration targets in the (top to bottom) blue, green, red, red edge, NIR bands for different exposure settings when gain was set to 1x. The circles represent the CRP calibrated estimated reflectance and the lines represent actual reflectance of the targets estimated with the handheld ASD FieldSpec-4 spectroradiometer. The red dash line and blue dotted line were used to mark the upper limits for full-scale (FS) multipoint ELM reflectance calibration and object-based (OB) reflectance calibration, respectively.**

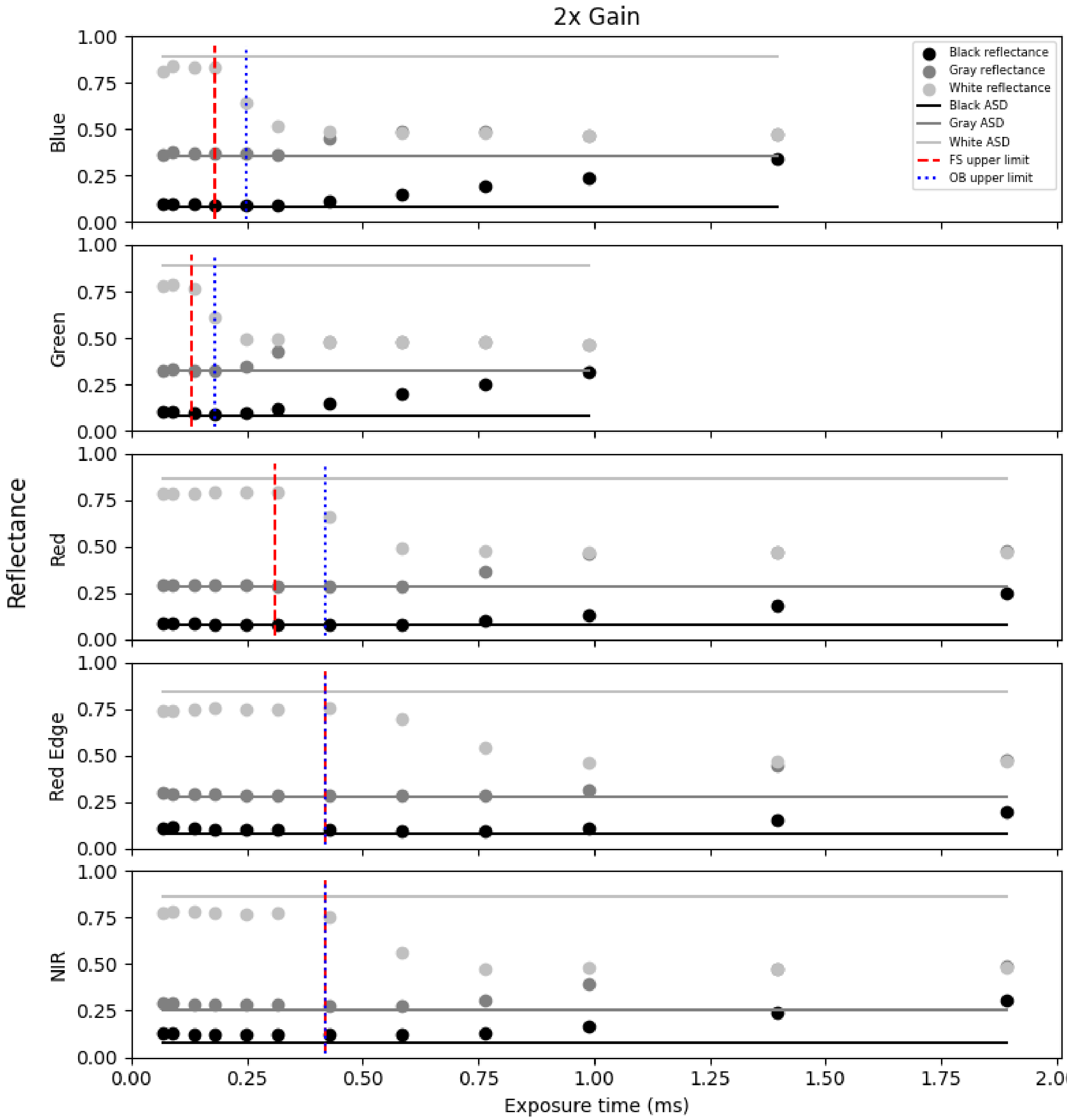

**Figure 5 Reflectance estimation of the in-field reflectance calibration targets in the (top to bottom) blue, green, red, red edge, NIR bands for different exposure settings when gain was set to 12x. The circles represent the CRP calibrated estimated reflectance and the lines represent actual reflectance of the black, gray, and white targets estimated with the handheld ASD FieldSpec-4 spectroradiometer. The red dash line and blue dotted line were used to mark the upper limits for full-scale (FS) multipoint ELM reflectance calibration and object-based (OB) reflectance calibration, respectively.**

**Table 5 Ideal exposure time for Micasense RedEdge-3 multispectral camera at 30 m AGL with gains 1x and 2x for capturing reflectance of all objects without distortion (full-scale reflectance) and capturing canopy and soil objects without distortion.**

| Bands | Ideal exp. time for full-scale reflectance (ms) | | Ideal exp. time for object-based (canopy and soil) reflectance (ms) | |
|---|---|---|---|---|
| | Gain 1x | Gain 2x | Gain 1x | Gain 2x |
| Blue | 0.25 – 0.43 | 0.09 – 0.18 | 0.25 – 0.59 | 0.09 – 0.31 |
| Green | 0.18 – 0.31 | 0.09 – 0.14 | 0.18 – 0.43 | 0.09 – 0.18 |
| Red | 0.31 – 0.77 | 0.18 – 0.31 | 0.31 – 0.99 | 0.18 – 0.42 |
| Red edge | 0.43– 0.99 | 0.14 – 0.43 | 0.43 – 0.99 | 0.18 – 0.43 |
| NIR | 0.43 – 0. 77 | 0.18 – 0.42 | 0.42 – 0.77 | 0.18 – 0.43 |

### 3.3 Cross-calibration accuracy matrix

To quantify the errors associated with variations in exposure time and gain during auto-exposure flights, cross-calibration was performed. Figure 6 through Figure 10 show the cross-calibration MAPE matrices for the five bands (in order of NIR, red edge, blue, green, and red) when object-based ELM from the reference images (vertical axis) was applied to target images (horizontal axis). All images in the top left and bottom right quadrants had 1x and 2x gains, respectively. The top right and bottom left quadrants represented scenarios where the reference and target gains were different. The color gradient from green to red in the matrices denote low to high MAPE from cross-calibration, respectively. The cells on the diagonals (marked by dotted black boxes) of the top left and bottom right quadrants, where the target and reference images had the same exposure and gain, had < 1 % MAPE in the visible and NIR bands. However, the diagonal cells in the red edge band had > 5 % errors, possibly from using all three calibration targets to derive ELM, resulting in a wider calibration range and greater errors compared to other bands. Additionally, larger solid black boxes were drawn to highlight the camera settings for which cross-calibration MAPE were ≤ 5 % for the visible and NIR bands, and ≤ 10 % for the red edge band. The exposure upper limit for the reference and target images within the back boxes corresponded to the ideal exposure upper limits (Table 5, Object-based ELM) for the respective gains. The box upper limits in the top right and bottom left quadrants of the blue (Figure 8) and green (Figure 9) bands did not coincide with the ideal upper limit but still were in the immediate neighborhood (±1 cell) of the ideal upper limit. For exposure values exceeding the upper limits, MAPE increased exponentially. An example can be seen with the NIR band (Figure 6), where the ideal exposure upper limit for 1x gain and 2x gain were determined to be 0.77 ms and 0.43 ms, respectively. Within the same figure, cell A (solid red box), for which the reference and target settings were within the ideal range, had 2.82 % error, as compared to cell B (solid cyan box), where the target image exposure time was beyond the ideal range, shows 19.87 % error. Likewise, cell C (maroon box), where the reference and target gains were 1x and 2x, respectively, and the exposure times were within limits, the error was 1.45 %; whereas cell D (blue box) had 27.84 % error as the reference image exposure was outside the upper limit for the NIR bands. While errors rates grew exponentially when exposure time increased, it was observed that the MAPE in most cases was within 10 – 15 % in the cells adjacent to the upper limits set by the bounding box.

Target images

| Reference image | Exposure time (ms) | 1x Gain | | | | | | | | | | | | | | | 2x Gain | | | | | | | | | | | | |
|---|---|---|---|---|---|---|---|---|---|---|---|---|---|---|---|---|---|---|---|---|---|---|---|---|---|---|---|---|---|
| | | 0.07 | 0.09 | 0.14 | 0.18 | 0.25 | 0.31 | 0.43 | 0.59 | 0.77 | 0.99 | 1.39 | 1.89 | 2.50 | 3.29 | 4.39 | 0.07 | 0.09 | 0.14 | 0.18 | 0.25 | 0.31 | 0.43 | 0.59 | 0.77 | 0.99 | 1.39 | 1.89 | 4.39 |
| 1x Gain | 0.07 | 1.04 | 0.58 | 1.06 | 1.19 | 1.97 | 2.75 | 4.99 | 2.94 | 2.05 | 5.35 | 19.18 | 41.09 | 70.26 | 72.69 | 72.29 | 6.66 | 6.59 | 5.19 | 4.63 | 3.74 | 3.23 | 1.33 | 14.74 | 27.73 | 50.73 | 72.49 | 75.03 | 126.72 |
| | 0.09 | 1.03 | 0.55 | 1.14 | 1.36 | 2.04 | 2.83 | 5.08 | 3.02 | 2.13 | 5.15 | 19.05 | 41.06 | 70.38 | 72.82 | 72.41 | 6.76 | 6.69 | 5.29 | 4.72 | 3.83 | 3.31 | 1.41 | 14.58 | 27.64 | 50.75 | 72.62 | 75.17 | 127.27 |
| | 0.14 | 1.49 | 1.24 | 0.39 | 1.01 | 1.22 | 1.78 | 3.88 | 1.86 | 1.32 | 5.40 | 19.51 | 40.69 | 69.50 | 71.89 | 71.49 | 5.53 | 5.46 | 4.08 | 3.53 | 2.65 | 2.14 | 0.46 | 14.67 | 27.50 | 50.21 | 71.70 | 74.20 | 126.17 |
| | 0.18 | 1.50 | 1.34 | 0.94 | 0.64 | 1.43 | 2.20 | 4.39 | 2.38 | 1.51 | 6.24 | 19.75 | 41.14 | 69.63 | 72.00 | 71.61 | 6.01 | 5.95 | 4.58 | 4.03 | 3.16 | 2.67 | 1.19 | 15.41 | 28.10 | 50.56 | 71.81 | 74.29 | 124.23 |
| | 0.25 | 2.30 | 2.12 | 1.20 | 1.40 | 0.25 | 0.79 | 2.90 | 0.94 | 0.24 | 6.34 | 19.84 | 40.62 | 68.59 | 70.92 | 70.53 | 4.50 | 4.44 | 3.10 | 2.56 | 1.71 | 1.21 | 1.35 | 15.35 | 27.81 | 49.86 | 70.73 | 73.16 | 123.40 |
| | 0.31 | 2.87 | 2.72 | 1.66 | 2.13 | 0.74 | 0.36 | 2.13 | 0.32 | 0.67 | 6.43 | 20.36 | 40.35 | 68.05 | 70.35 | 69.97 | 3.71 | 3.65 | 2.32 | 1.79 | 1.26 | 0.82 | 1.50 | 15.39 | 27.67 | 49.51 | 70.16 | 72.57 | 122.90 |
| | 0.43 | 4.59 | 4.67 | 3.63 | 4.14 | 2.79 | 2.07 | 0.24 | 1.89 | 2.72 | 8.23 | 21.88 | 39.61 | 66.55 | 68.79 | 68.42 | 2.63 | 2.64 | 1.72 | 1.50 | 1.38 | 1.63 | 3.38 | 16.90 | 27.28 | 48.52 | 68.61 | 70.95 | 121.63 |
| | 0.59 | 3.05 | 2.90 | 1.86 | 2.30 | 0.92 | 0.49 | 1.93 | 0.18 | 0.86 | 6.64 | 20.41 | 40.35 | 67.86 | 70.15 | 69.77 | 3.51 | 3.45 | 2.18 | 1.74 | 1.34 | 1.06 | 1.74 | 15.52 | 27.75 | 49.44 | 69.96 | 72.36 | 122.28 |
| | 0.77 | 2.38 | 2.20 | 1.30 | 1.47 | 0.29 | 0.70 | A) 2.82 | 0.85 | 0.25 | 6.43 | B) 19.87 | 40.61 | 68.51 | 70.84 | 70.45 | 4.41 | 4.35 | 3.01 | 2.47 | 1.66 | 1.15 | C) 1.45 | 15.41 | D) 27.84 | 49.83 | 70.64 | 73.07 | 123.14 |
| | 0.99 | 6.47 | 5.96 | 6.36 | 7.38 | 7.65 | 8.13 | 10.33 | 8.20 | 7.78 | 0.30 | 17.12 | 41.43 | 75.23 | 78.04 | 77.57 | 12.26 | 12.19 | 10.57 | 9.92 | 8.88 | 8.29 | 6.16 | 11.00 | 25.95 | 52.60 | 77.81 | 80.75 | 142.27 |
| | 1.39 | 42.57 | 41.54 | 42.87 | 43.70 | 44.85 | 46.42 | 51.27 | 46.83 | 44.97 | 31.99 | 0.64 | 47.43 | 110.54 | 115.79 | 114.91 | 54.87 | 54.73 | 51.71 | 50.49 | 48.56 | 47.46 | 43.36 | 11.67 | 18.53 | 68.28 | 115.35 | 120.85 | 233.65 |
| | 1.89 | 173.39 | 172.54 | 174.00 | 177.88 | 178.86 | 179.50 | 181.12 | 180.66 | 179.33 | 150.93 | 92.52 | 0.61 | 125.21 | 137.99 | 134.96 | 169.11 | 169.04 | 173.28 | 174.27 | 174.87 | 175.91 | 172.92 | 111.29 | 56.40 | 45.48 | 137.53 | 154.43 | 530.28 |
| | 2.50 | 2841.5 | 2833.5 | 2847.1 | 2883.3 | 2892.5 | 2898.4 | 2913.5 | 2909.2 | 2896.8 | 2632.1 | 2087.5 | 1225.1 | 76.56 | 77.28 | 48.96 | 2801.6 | 2800.9 | 2840.4 | 2849.6 | 2855.3 | 2864.9 | 2837.1 | 2262.5 | 1750.9 | 845.5 | 73.0 | 230.5 | 6152.3 |
| | 3.29 | 4433.9 | 4421.5 | 4442.6 | 4498.6 | 4512.8 | 4521.9 | 4545.3 | 4538.7 | 4519.5 | 4110.0 | 3267.5 | 1994.7 | 189.0 | 14.8 | 48.5 | 4372.1 | 4371.1 | 4432.2 | 4446.5 | 4455.2 | 4470.1 | 4427.1 | 3538.2 | 2746.7 | 1353.7 | 64.0 | 232.2 | 9642.1 |
| | 4.39 | 4452.6 | 4440.2 | 4461.5 | 4518.0 | 4532.3 | 4541.5 | 4565.2 | 4558.5 | 4539.1 | 4125.6 | 3274.9 | 1970.1 | 146.7 | 40.0 | 24.8 | 4390.3 | 4389.2 | 4451.0 | 4465.4 | 4474.2 | 4489.3 | 4445.8 | 3548.2 | 2749.0 | 1334.5 | 62.9 | 278.6 | 9692.4 |
| 2x Gain | 0.07 | 6.33 | 6.41 | 5.33 | 5.86 | 4.47 | 3.72 | 2.70 | 3.54 | 4.39 | 10.09 | 24.20 | 38.24 | 66.09 | 68.40 | 68.02 | 0.88 | 0.44 | 1.39 | 1.93 | 2.78 | 3.26 | 5.07 | 19.05 | 25.49 | 47.44 | 68.21 | 70.64 | 127.30 |
| | 0.09 | 6.27 | 6.35 | 5.28 | 5.81 | 4.41 | 3.66 | 2.72 | 3.48 | 4.33 | 10.04 | 24.16 | 38.26 | 66.13 | 68.45 | 68.06 | 0.89 | 0.44 | 1.35 | 1.87 | 2.72 | 3.21 | 5.02 | 19.01 | 25.50 | 47.47 | 68.26 | 70.68 | 127.36 |
| | 0.14 | 4.92 | 5.00 | 3.93 | 4.46 | 3.07 | 2.32 | 1.76 | 2.14 | 2.99 | 8.67 | 22.73 | 39.05 | 66.81 | 69.12 | 68.73 | 1.49 | 1.42 | 0.35 | 0.56 | 1.38 | 1.87 | 3.67 | 17.61 | 26.34 | 48.22 | 68.93 | 71.34 | 125.51 |
| | 0.18 | 4.39 | 4.47 | 3.40 | 3.93 | 2.54 | 1.79 | 1.54 | 1.63 | 2.46 | 8.14 | 22.22 | 39.32 | 67.12 | 69.43 | 69.04 | 1.97 | 1.87 | 0.65 | 0.43 | 0.85 | 1.33 | 3.14 | 17.09 | 26.60 | 48.51 | 69.24 | 71.66 | 125.14 |
| | 0.25 | 3.56 | 3.64 | 2.56 | 3.09 | 1.69 | 1.08 | 1.47 | 1.30 | 1.62 | 7.34 | 21.50 | 39.70 | 67.66 | 69.99 | 69.60 | 2.80 | 2.74 | 1.40 | 0.86 | 0.34 | 0.51 | 2.30 | 16.34 | 26.90 | 48.94 | 69.79 | 72.23 | 125.03 |
| | 0.31 | 3.11 | 3.16 | 2.08 | 2.61 | 1.21 | 0.81 | 1.69 | 1.08 | 1.14 | 6.86 | 21.03 | 39.95 | 67.93 | 70.26 | 69.87 | 3.29 | 3.23 | 1.89 | 1.35 | 0.66 | 0.19 | 1.82 | 15.86 | 27.14 | 49.20 | 70.07 | 72.51 | 124.63 |
| | 0.43 | 1.66 | 1.41 | 0.47 | 1.27 | 1.39 | 1.77 | 3.62 | 1.85 | 1.50 | 5.20 | 19.82 | 40.52 | 69.38 | 71.78 | 71.38 | 5.27 | 5.21 | 3.82 | 3.27 | 2.38 | 1.88 | 0.27 | 14.57 | 27.30 | 50.06 | 71.58 | 74.09 | 126.68 |
| | 0.59 | 25.09 | 24.35 | 24.92 | 26.40 | 26.79 | 27.47 | 30.65 | 27.58 | 26.98 | 15.77 | 9.02 | 44.12 | 92.97 | 97.03 | 96.36 | 33.44 | 33.33 | 30.99 | 30.04 | 28.55 | 27.70 | 24.62 | 0.21 | 21.76 | 60.27 | 96.70 | 100.95 | 189.87 |
| | 0.77 | 76.36 | 75.80 | 76.76 | 79.29 | 79.93 | 80.35 | 81.41 | 81.11 | 80.24 | 61.70 | 23.55 | 36.87 | 117.32 | 124.02 | 122.90 | 73.57 | 73.52 | 76.29 | 76.93 | 77.33 | 78.00 | 76.06 | 35.81 | 0.67 | 63.75 | 123.67 | 133.58 | 313.54 |
| | 0.99 | 300.43 | 299.23 | 301.28 | 306.74 | 308.12 | 309.01 | 311.29 | 310.65 | 308.77 | 268.90 | 186.86 | 63.72 | 116.09 | 130.72 | 128.09 | 294.42 | 294.32 | 300.27 | 301.67 | 302.52 | 303.97 | 299.78 | 213.22 | 136.25 | 2.09 | 130.27 | 153.13 | 808.42 |
| | 1.39 | 2713.3 | 2705.9 | 2718.6 | 2752.5 | 2761.1 | 2766.6 | 2780.8 | 2776.8 | 2765.1 | 2517.4 | 2007.7 | 1223.7 | 131.5 | 36.0 | 51.5 | 2676.0 | 2675.4 | 2712.4 | 2721.0 | 2726.3 | 2735.3 | 2709.3 | 2171.5 | 1692.6 | 845.1 | 52.7 | 123.7 | 5850.4 |
| | 1.89 | 2828.2 | 2820.6 | 2833.6 | 2868.0 | 2876.7 | 2882.3 | 2896.6 | 2892.6 | 2880.8 | 2629.4 | 2112.2 | 1380.3 | 271.8 | 158.6 | 185.5 | 2790.3 | 2789.7 | 2827.2 | 2836.0 | 2841.3 | 2850.5 | 2824.1 | 2278.4 | 1824.0 | 977.7 | 162.7 | 35.2 | 6075.1 |
| | 4.39 | 77.96 | 77.96 | 77.96 | 77.96 | 77.96 | 77.96 | 77.96 | 77.96 | 77.96 | 77.96 | 77.96 | 77.96 | 77.96 | 77.96 | 77.96 | 77.96 | 77.96 | 77.96 | 77.96 | 77.96 | 77.96 | 77.96 | 77.96 | 77.96 | 77.96 | 77.96 | 77.96 | 77.96 |

**Figure 6 Object-based cross-calibration MAPE for the NIR band for various exposure and gain combinations, where the acceptable range is marked by the back box.**

Target images

| Reference image | Exposure time (ms) | 1x Gain | | | | | | | | | | | | | | | 2x Gain | | | | | | | | | | | | |
|---|---|---|---|---|---|---|---|---|---|---|---|---|---|---|---|---|---|---|---|---|---|---|---|---|---|---|---|---|---|
| | | 0.068 | 0.090 | 0.135 | 0.180 | 0.248 | 0.315 | 0.428 | 0.585 | 0.765 | 0.990 | 1.395 | 1.890 | 2.500 | 3.289 | 4.386 | 0.068 | 0.090 | 0.135 | 0.180 | 0.248 | 0.315 | 0.428 | 0.585 | 0.765 | 0.990 | 1.395 | 1.890 | 4.386 |
| 1x Gain | 0.068 | 154.75 | 173.55 | 173.39 | 172.35 | 171.71 | 171.61 | 171.06 | 170.98 | 170.79 | 170.85 | 173.63 | 177.68 | 188.87 | 201.62 | 212.59 | 175.22 | 175.19 | 173.60 | 172.93 | 172.46 | 171.88 | 171.73 | 172.44 | 175.07 | 180.63 | 196.60 | 207.79 | 150.19 |
| | 0.090 | 110.07 | 7.51 | 7.45 | 10.34 | 11.46 | 12.46 | 13.90 | 13.43 | 13.19 | 13.82 | 18.79 | 25.53 | 37.08 | 70.37 | 98.97 | 4.35 | 4.44 | 7.94 | 9.27 | 9.40 | 10.74 | 11.06 | 12.32 | 20.77 | 23.45 | 57.26 | 86.44 | 135.31 |
| | 0.135 | 109.03 | 7.11 | 6.96 | 9.61 | 10.73 | 11.72 | 13.15 | 12.68 | 12.45 | 13.07 | 18.31 | 25.01 | 37.46 | 70.53 | 98.95 | 3.82 | 3.86 | 7.23 | 8.56 | 8.68 | 10.01 | 10.33 | 11.89 | 20.28 | 22.95 | 57.51 | 86.50 | 134.11 |
| | 0.180 | 104.10 | 5.41 | 5.23 | 6.39 | 7.58 | 8.08 | 9.46 | 9.01 | 8.91 | 9.38 | 16.21 | 22.67 | 38.81 | 70.70 | 98.10 | 4.70 | 4.68 | 4.13 | 5.11 | 5.89 | 6.84 | 6.98 | 10.02 | 18.11 | 20.68 | 58.15 | 86.10 | 128.38 |
| | 0.248 | 102.40 | 4.49 | 4.22 | 5.41 | 6.57 | 7.18 | 8.55 | 8.10 | 7.96 | 8.47 | 15.18 | 21.63 | 40.23 | 72.05 | 99.39 | 6.19 | 6.11 | 3.40 | 4.19 | 4.89 | 5.83 | 5.99 | 9.00 | 17.08 | 20.20 | 59.53 | 87.42 | 126.51 |
| | 0.315 | 101.24 | 4.47 | 4.15 | 5.29 | 6.45 | 6.39 | 7.40 | 7.25 | 7.66 | 7.70 | 14.92 | 21.26 | 39.82 | 71.10 | 97.98 | 6.36 | 6.28 | 3.41 | 3.77 | 4.80 | 5.73 | 5.87 | 8.85 | 16.79 | 19.96 | 58.79 | 86.20 | 125.09 |
| | 0.428 | 98.84 | 4.07 | 3.89 | 4.39 | 5.54 | 5.48 | 5.96 | 6.30 | 6.72 | 6.76 | 13.86 | 20.09 | 40.59 | 71.33 | 97.75 | 7.70 | 7.62 | 3.82 | 2.93 | 3.92 | 4.83 | 4.96 | 7.89 | 15.70 | 20.74 | 59.23 | 86.18 | 122.31 |
| | 0.585 | 99.26 | 4.04 | 3.93 | 4.21 | 5.37 | 5.31 | 6.26 | 6.15 | 6.57 | 6.61 | 13.80 | 20.10 | 41.24 | 72.36 | 99.10 | 7.95 | 7.87 | 4.01 | 2.81 | 3.73 | 4.65 | 4.79 | 7.75 | 15.65 | 21.15 | 60.11 | 87.39 | 122.88 |
| | 0.765 | 99.24 | 4.37 | 4.19 | 3.84 | 5.02 | 5.21 | 6.57 | 6.12 | 6.23 | 6.49 | 13.52 | 19.88 | 42.09 | 73.50 | 100.50 | 8.48 | 8.39 | 4.50 | 2.97 | 3.36 | 4.29 | 4.43 | 7.42 | 15.40 | 21.80 | 61.14 | 88.67 | 122.95 |
| | 0.990 | 99.12 | 4.13 | 4.00 | 4.08 | 5.25 | 5.18 | 6.22 | 6.04 | 6.45 | 6.49 | 13.68 | 19.99 | 41.46 | 72.61 | 99.38 | 8.14 | 8.06 | 4.19 | 2.83 | 3.60 | 4.53 | 4.66 | 7.63 | 15.54 | 21.35 | 60.35 | 87.66 | 122.74 |
| | 1.395 | 131.18 | 23.00 | 23.19 | 27.01 | 28.39 | 29.64 | 31.42 | 30.83 | 30.54 | 31.31 | 23.57 | 31.91 | 41.62 | 82.79 | 118.17 | 19.58 | 19.71 | 24.05 | 25.69 | 25.85 | 27.50 | 27.90 | 25.08 | 26.03 | 29.34 | 66.59 | 102.68 | 157.72 |
| | 1.890 | 177.14 | 55.35 | 55.59 | 60.70 | 62.55 | 64.21 | 66.59 | 65.80 | 65.41 | 66.44 | 53.75 | 48.91 | 45.79 | 96.00 | 143.26 | 50.77 | 50.95 | 56.74 | 58.94 | 59.15 | 61.36 | 61.89 | 58.13 | 46.26 | 45.48 | 74.35 | 122.56 | 206.90 |
| | 2.500 | 189.78 | 58.72 | 58.94 | 63.62 | 65.32 | 66.84 | 69.02 | 68.30 | 67.94 | 68.88 | 57.26 | 61.66 | 58.80 | 64.44 | 107.74 | 54.53 | 54.69 | 60.00 | 62.01 | 62.21 | 64.22 | 64.71 | 61.27 | 54.46 | 58.52 | 53.47 | 88.77 | 221.92 |
| | 3.289 | 173.50 | 46.47 | 46.96 | 50.43 | 52.52 | 52.86 | 54.71 | 54.96 | 55.58 | 55.38 | 52.30 | 58.54 | 56.14 | 46.69 | 83.05 | 41.69 | 41.69 | 46.27 | 48.48 | 50.04 | 51.96 | 52.48 | 50.10 | 55.82 | 55.91 | 51.66 | 67.13 | 204.82 |
| | 4.386 | 264.30 | 112.32 | 112.98 | 117.24 | 119.82 | 120.24 | 122.51 | 122.82 | 123.58 | 123.33 | 114.03 | 118.63 | 104.04 | 89.75 | 47.46 | 105.51 | 105.64 | 112.10 | 114.84 | 116.76 | 119.13 | 119.77 | 116.84 | 118.36 | 106.43 | 98.53 | 64.62 | 301.74 |
| 2x Gain | 0.068 | 114.17 | 10.24 | 10.06 | 12.26 | 13.37 | 14.37 | 15.81 | 15.34 | 15.10 | 15.72 | 21.49 | 28.21 | 32.95 | 66.14 | 94.66 | 6.65 | 6.38 | 9.87 | 11.19 | 11.32 | 12.65 | 12.97 | 15.04 | 23.47 | 26.14 | 53.08 | 82.17 | 139.75 |
| | 0.090 | 113.31 | 10.07 | 9.88 | 11.54 | 12.65 | 13.64 | 15.06 | 14.60 | 14.36 | 14.98 | 21.21 | 27.87 | 32.92 | 65.81 | 94.07 | 6.48 | 6.05 | 9.18 | 10.49 | 10.72 | 11.94 | 12.25 | 14.82 | 23.17 | 25.82 | 52.86 | 81.69 | 138.72 |
| | 0.135 | 107.80 | 7.44 | 7.26 | 8.44 | 9.63 | 10.19 | 11.58 | 11.12 | 11.00 | 11.49 | 18.33 | 24.85 | 36.09 | 68.25 | 95.88 | 3.93 | 3.49 | 6.16 | 7.17 | 7.93 | 8.88 | 9.03 | 12.09 | 20.25 | 22.84 | 55.59 | 83.78 | 132.51 |
| | 0.180 | 105.14 | 6.26 | 6.08 | 7.24 | 8.42 | 8.46 | 9.81 | 9.40 | 9.64 | 9.77 | 17.01 | 23.44 | 37.41 | 69.14 | 96.41 | 3.80 | 3.86 | 4.99 | 5.70 | 6.74 | 7.69 | 7.82 | 10.85 | 18.90 | 21.46 | 56.65 | 84.47 | 129.48 |
| | 0.248 | 104.60 | 5.44 | 5.26 | 6.61 | 7.67 | 8.57 | 9.96 | 9.50 | 9.27 | 9.87 | 16.32 | 22.83 | 38.99 | 71.11 | 98.71 | 4.63 | 4.63 | 4.30 | 5.49 | 5.99 | 6.93 | 7.21 | 10.09 | 18.24 | 20.83 | 58.47 | 86.62 | 129.00 |
| | 0.315 | 102.63 | 4.56 | 4.36 | 5.54 | 6.70 | 7.30 | 8.68 | 8.23 | 8.09 | 8.60 | 15.32 | 21.77 | 40.04 | 71.87 | 99.22 | 5.99 | 5.91 | 3.47 | 4.31 | 5.02 | 5.97 | 6.13 | 9.14 | 17.22 | 20.21 | 59.34 | 87.25 | 126.77 |
| | 0.428 | 101.97 | 4.43 | 4.08 | 5.24 | 6.42 | 6.86 | 8.23 | 7.78 | 7.71 | 8.15 | 15.01 | 21.43 | 40.33 | 72.05 | 99.29 | 6.41 | 6.32 | 3.37 | 3.91 | 4.75 | 5.69 | 5.83 | 8.85 | 16.90 | 20.15 | 59.56 | 87.36 | 126.02 |
| | 0.585 | 111.22 | 8.70 | 8.85 | 12.08 | 13.25 | 14.30 | 15.81 | 15.31 | 15.06 | 15.71 | 17.86 | 24.90 | 40.64 | 75.42 | 105.31 | 7.25 | 7.90 | 9.58 | 10.97 | 11.10 | 12.50 | 12.83 | 11.10 | 19.93 | 22.73 | 61.73 | 92.22 | 136.42 |
| | 0.765 | 150.92 | 37.11 | 37.33 | 41.73 | 43.33 | 44.76 | 46.81 | 46.14 | 45.79 | 46.68 | 35.74 | 38.89 | 42.41 | 89.86 | 130.62 | 33.17 | 33.32 | 38.32 | 40.21 | 40.40 | 42.30 | 42.75 | 39.51 | 32.12 | 35.94 | 71.18 | 112.77 | 178.78 |
| | 0.990 | 188.42 | 61.15 | 61.40 | 66.52 | 68.37 | 70.03 | 72.42 | 71.64 | 71.24 | 72.27 | 59.55 | 56.20 | 53.07 | 85.65 | 133.03 | 56.56 | 56.75 | 62.55 | 64.75 | 64.97 | 67.18 | 67.71 | 63.94 | 52.04 | 52.76 | 63.95 | 112.28 | 219.55 |
| | 1.395 | 176.17 | 47.94 | 48.14 | 52.27 | 53.76 | 55.11 | 57.03 | 56.40 | 56.08 | 56.91 | 49.37 | 58.38 | 55.85 | 52.89 | 91.10 | 44.24 | 44.39 | 49.07 | 50.85 | 51.02 | 52.80 | 53.23 | 50.19 | 52.02 | 55.61 | 51.15 | 74.37 | 207.68 |
| | 1.890 | 218.42 | 79.21 | 79.80 | 83.65 | 85.98 | 86.36 | 88.41 | 88.69 | 89.38 | 89.15 | 83.23 | 87.39 | 79.92 | 67.02 | 64.69 | 73.06 | 73.18 | 79.01 | 81.48 | 83.22 | 85.36 | 85.94 | 83.29 | 87.14 | 79.66 | 74.94 | 46.99 | 252.72 |
| | 4.386 | 167.93 | 167.93 | 167.93 | 167.93 | 167.93 | 167.93 | 167.93 | 167.93 | 167.93 | 167.93 | 167.93 | 167.93 | 167.93 | 167.93 | 167.93 | 167.93 | 167.93 | 167.93 | 167.93 | 167.93 | 167.93 | 167.93 | 167.93 | 167.93 | 167.93 | 167.93 | 167.93 | 167.93 |

**Figure 7 Object-based cross-calibration MAPE for the red edge band for various exposure and gain combinations, where the acceptable range is marked by the back box.**

Target images

| Reference image | Exposure time (ms) | 1x Gain | | | | | | | | | | | | | | | 2x Gain | | | | | | | | | | | | |
|---|---|---|---|---|---|---|---|---|---|---|---|---|---|---|---|---|---|---|---|---|---|---|---|---|---|---|---|---|---|
| | | 0.068 | 0.090 | 0.135 | 0.180 | 0.248 | 0.315 | 0.428 | 0.585 | 0.765 | 0.990 | 1.395 | 1.890 | 2.500 | 3.289 | 4.386 | 0.068 | 0.090 | 0.135 | 0.180 | 0.248 | 0.315 | 0.428 | 0.585 | 0.765 | 0.990 | 1.395 | 1.890 | 4.386 |
| 1x Gain | 0.068 | 0.98 | 3.01 | 3.43 | 5.43 | 5.06 | 5.00 | 4.98 | 5.46 | 5.19 | 31.60 | 60.77 | 94.37 | 136.98 | 192.89 | 221.66 | 3.89 | 5.44 | 2.17 | 3.60 | 3.57 | 4.22 | 25.01 | 51.33 | 77.28 | 104.43 | 165.57 | 216.03 | 225.24 |
| | 0.090 | 3.04 | 0.41 | 0.98 | 2.52 | 3.05 | 3.50 | 3.48 | 3.96 | 4.46 | 34.81 | 64.15 | 97.93 | 140.78 | 197.00 | 224.71 | 6.95 | 8.50 | 5.02 | 3.38 | 2.86 | 2.72 | 28.18 | 54.65 | 80.74 | 108.05 | 169.52 | 219.05 | 228.31 |
| | 0.135 | 3.44 | 0.98 | 0.28 | 1.99 | 2.05 | 2.48 | 2.46 | 2.94 | 4.64 | 34.81 | 63.81 | 97.20 | 139.56 | 195.12 | 222.35 | 7.28 | 8.82 | 5.38 | 3.76 | 2.86 | 2.00 | 28.27 | 54.43 | 80.22 | 107.20 | 167.97 | 216.75 | 225.91 |
| | 0.180 | 5.39 | 2.52 | 1.98 | 0.42 | 0.93 | 1.32 | 3.45 | 2.07 | 6.61 | 36.69 | 65.60 | 98.89 | 141.12 | 196.52 | 222.87 | 9.24 | 10.78 | 7.35 | 5.73 | 4.84 | 3.81 | 30.16 | 56.25 | 81.96 | 108.86 | 169.44 | 217.29 | 226.42 |
| | 0.248 | 5.06 | 2.97 | 2.02 | 0.84 | 0.52 | 0.94 | 3.04 | 1.63 | 6.16 | 35.83 | 64.34 | 97.17 | 138.81 | 193.45 | 219.58 | 8.75 | 10.27 | 6.88 | 5.29 | 4.41 | 3.40 | 29.39 | 55.11 | 80.47 | 107.00 | 166.75 | 214.08 | 223.08 |
| | 0.315 | 4.82 | 3.35 | 2.41 | 1.46 | 1.07 | 0.16 | 2.07 | 0.83 | 5.14 | 34.33 | 62.38 | 94.69 | 135.66 | 189.42 | 215.50 | 7.69 | 9.18 | 5.85 | 4.28 | 3.42 | 2.42 | 27.99 | 53.30 | 78.25 | 104.36 | 163.15 | 210.09 | 218.95 |
| | 0.428 | 4.71 | 3.27 | 2.35 | 3.27 | 2.92 | 2.02 | 0.30 | 1.38 | 3.01 | 31.62 | 59.11 | 90.77 | 130.93 | 183.61 | 209.98 | 5.95 | 6.97 | 3.71 | 2.17 | 1.41 | 0.75 | 25.41 | 50.22 | 74.67 | 100.25 | 157.86 | 204.68 | 213.35 |
| | 0.585 | 5.19 | 3.74 | 2.81 | 2.01 | 1.64 | 0.69 | 1.39 | 0.38 | 4.42 | 33.21 | 60.89 | 92.75 | 133.17 | 186.19 | 212.18 | 6.94 | 8.41 | 5.12 | 3.57 | 2.72 | 1.74 | 26.96 | 51.93 | 76.54 | 102.29 | 160.28 | 206.84 | 215.58 |
| | 0.765 | 4.75 | 3.92 | 4.25 | 6.08 | 5.74 | 4.87 | 2.91 | 4.24 | 0.37 | 27.69 | 54.30 | 84.94 | 123.80 | 174.78 | 201.47 | 5.84 | 5.50 | 3.27 | 1.58 | 2.06 | 2.57 | 21.68 | 45.69 | 69.35 | 94.11 | 149.87 | 196.34 | 204.74 |
| | 0.990 | 23.12 | 25.34 | 25.64 | 27.11 | 26.84 | 26.14 | 24.57 | 25.64 | 22.23 | 0.32 | 21.36 | 46.72 | 77.88 | 119.62 | 148.67 | 20.28 | 19.15 | 21.68 | 22.88 | 23.54 | 24.29 | 4.82 | 14.45 | 33.45 | 54.09 | 98.17 | 145.00 | 151.21 |
| | 1.395 | 51.00 | 53.54 | 53.89 | 55.57 | 55.26 | 54.46 | 52.65 | 53.88 | 49.97 | 24.49 | 0.69 | 29.44 | 65.16 | 112.98 | 156.05 | 47.75 | 46.44 | 49.35 | 50.72 | 51.47 | 52.34 | 30.02 | 10.03 | 13.86 | 37.89 | 88.33 | 151.84 | 158.81 |
| | 1.890 | 98.46 | 101.62 | 102.06 | 104.14 | 103.76 | 102.76 | 100.52 | 102.04 | 97.19 | 66.58 | 36.59 | 0.29 | 44.68 | 104.66 | 172.22 | 94.41 | 92.80 | 96.41 | 98.11 | 99.05 | 100.13 | 72.38 | 49.07 | 21.94 | 10.91 | 74.30 | 166.71 | 175.99 |
| | 2.500 | 197.26 | 201.61 | 202.21 | 205.09 | 204.56 | 203.18 | 200.09 | 202.20 | 195.49 | 153.15 | 111.75 | 61.42 | 2.38 | 83.14 | 200.92 | 191.67 | 189.43 | 194.42 | 196.77 | 198.07 | 199.55 | 161.25 | 128.98 | 91.53 | 47.36 | 41.21 | 193.34 | 206.13 |
| | 3.289 | 551.38 | 560.08 | 561.28 | 567.03 | 565.97 | 563.22 | 557.04 | 561.25 | 547.86 | 464.31 | 381.65 | 280.81 | 158.57 | 11.86 | 306.66 | 540.22 | 535.76 | 545.72 | 550.42 | 553.00 | 555.97 | 479.49 | 416.05 | 341.28 | 252.68 | 79.04 | 291.13 | 316.96 |
| | 4.386 | 99.16 | 99.16 | 99.16 | 99.15 | 99.15 | 99.15 | 99.15 | 99.15 | 99.15 | 99.15 | 99.19 | 99.24 | 99.30 | 99.37 | 99.39 | 99.16 | 99.16 | 99.16 | 99.15 | 99.15 | 99.15 | 99.15 | 99.18 | 99.21 | 99.25 | 99.33 | 99.38 | 99.39 |
| 2x Gain | 0.068 | 3.80 | 6.77 | 7.18 | 9.14 | 8.78 | 7.84 | 6.13 | 7.17 | 6.27 | 27.19 | 55.81 | 88.78 | 130.58 | 185.44 | 215.19 | 0.60 | 1.53 | 2.76 | 4.72 | 4.68 | 5.50 | 20.72 | 46.55 | 72.01 | 98.65 | 158.63 | 209.66 | 218.70 |
| | 0.090 | 5.21 | 8.12 | 8.52 | 10.44 | 10.09 | 9.17 | 7.10 | 8.51 | 5.78 | 25.13 | 53.16 | 85.44 | 126.38 | 180.10 | 209.83 | 1.49 | 0.33 | 3.32 | 4.89 | 5.75 | 6.75 | 18.80 | 44.09 | 69.02 | 95.11 | 153.85 | 204.42 | 213.27 |
| | 0.135 | 2.47 | 4.77 | 5.17 | 7.09 | 6.73 | 5.82 | 3.76 | 5.16 | 3.43 | 28.32 | 56.22 | 88.34 | 129.09 | 182.55 | 210.82 | 2.70 | 3.31 | 0.40 | 1.94 | 2.42 | 3.41 | 22.02 | 47.19 | 72.00 | 97.97 | 156.43 | 205.43 | 214.24 |
| | 0.180 | 3.41 | 3.16 | 3.55 | 5.44 | 5.09 | 4.19 | 2.17 | 3.54 | 1.59 | 29.43 | 56.91 | 88.55 | 128.67 | 181.32 | 208.55 | 4.54 | 4.80 | 1.89 | 0.59 | 1.08 | 1.82 | 23.23 | 48.01 | 72.45 | 98.02 | 155.59 | 203.24 | 211.92 |
| | 0.248 | 3.41 | 2.45 | 2.73 | 4.63 | 4.28 | 3.37 | 1.53 | 2.72 | 1.98 | 30.55 | 58.27 | 90.20 | 130.69 | 183.82 | 210.94 | 4.66 | 5.70 | 2.41 | 0.93 | 0.60 | 1.02 | 24.29 | 49.30 | 73.96 | 99.76 | 157.85 | 205.59 | 214.34 |
| | 0.315 | 4.03 | 2.58 | 1.86 | 3.65 | 3.30 | 2.39 | 0.76 | 1.74 | 2.68 | 31.52 | 59.24 | 91.15 | 131.63 | 184.74 | 211.46 | 5.37 | 6.68 | 3.39 | 1.84 | 0.99 | 0.33 | 25.26 | 50.27 | 74.92 | 100.71 | 158.78 | 206.12 | 214.86 |
| | 0.428 | 19.09 | 21.40 | 21.72 | 23.25 | 22.96 | 22.23 | 20.59 | 21.71 | 18.15 | 5.03 | 27.31 | 52.97 | 85.51 | 128.81 | 157.82 | 16.13 | 14.94 | 17.58 | 18.83 | 19.52 | 20.31 | 0.25 | 20.10 | 39.92 | 60.66 | 107.35 | 153.52 | 160.56 |
| | 0.585 | 39.76 | 42.10 | 42.42 | 43.97 | 43.69 | 42.94 | 41.28 | 42.41 | 38.81 | 15.29 | 9.27 | 36.43 | 69.40 | 113.53 | 149.96 | 36.75 | 35.55 | 38.23 | 39.50 | 40.19 | 40.99 | 20.39 | 0.19 | 20.30 | 44.23 | 90.78 | 146.07 | 152.50 |
| | 0.765 | 68.45 | 71.13 | 71.50 | 73.27 | 72.95 | 72.10 | 70.20 | 71.49 | 67.37 | 40.47 | 14.62 | 18.64 | 56.35 | 106.82 | 157.69 | 65.02 | 63.64 | 66.71 | 68.16 | 68.95 | 69.87 | 46.31 | 23.02 | 0.69 | 27.56 | 80.80 | 153.25 | 160.60 |
| | 0.990 | 116.54 | 119.91 | 120.38 | 122.61 | 122.20 | 121.13 | 118.73 | 120.37 | 115.17 | 82.42 | 50.34 | 11.31 | 36.59 | 100.74 | 177.49 | 112.21 | 110.48 | 114.34 | 116.16 | 117.17 | 118.32 | 88.64 | 63.69 | 34.68 | 0.93 | 68.26 | 171.59 | 181.52 |
| | 1.395 | 311.00 | 316.68 | 317.46 | 321.22 | 320.53 | 318.73 | 314.69 | 317.44 | 308.69 | 251.92 | 197.62 | 133.73 | 53.67 | 55.30 | 230.51 | 303.71 | 300.79 | 307.29 | 310.37 | 312.05 | 313.99 | 264.03 | 220.09 | 171.25 | 114.82 | 2.15 | 221.09 | 236.99 |
| | 1.890 | 97.10 | 97.08 | 97.07 | 97.05 | 97.04 | 97.03 | 97.03 | 97.02 | 97.03 | 97.06 | 97.45 | 97.92 | 98.50 | 99.27 | 99.42 | 97.12 | 97.11 | 97.08 | 97.05 | 97.05 | 97.04 | 97.05 | 97.29 | 97.64 | 98.06 | 98.88 | 99.35 | 99.46 |
| | 4.386 | 100.57 | 100.58 | 100.58 | 100.59 | 100.60 | 100.60 | 100.60 | 100.61 | 100.60 | 100.59 | 100.39 | 100.15 | 99.86 | 99.47 | 99.40 | 100.56 | 100.56 | 100.58 | 100.59 | 100.59 | 100.60 | 100.59 | 100.47 | 100.30 | 100.08 | 99.67 | 99.43 | 99.38 |

**Figure 8 Object-based cross-calibration MAPE for the blue band for various exposure and gain combinations, where the acceptable range is marked by the back box.**

Target images

| Reference image | Exposure time (ms) | 1x Gain | | | | | | | | | | | | | | | 2x Gain | | | | | | | | | | | | |
|---|---|---|---|---|---|---|---|---|---|---|---|---|---|---|---|---|---|---|---|---|---|---|---|---|---|---|---|---|---|
| | | 0.068 | 0.090 | 0.135 | 0.180 | 0.248 | 0.315 | 0.428 | 0.585 | 0.765 | 0.990 | 1.395 | 1.890 | 2.500 | 3.289 | 4.386 | 0.068 | 0.090 | 0.135 | 0.180 | 0.248 | 0.315 | 0.428 | 0.585 | 0.765 | 0.990 | 1.395 | 1.890 | 4.386 |
| 1x Gain | 0.068 | 0.70 | 2.68 | 4.55 | 5.97 | 7.21 | 8.46 | 9.52 | 9.67 | 39.34 | 65.21 | 105.41 | 152.23 | 212.70 | 249.59 | 250.07 | 121.65 | 4.61 | 2.08 | 3.55 | 6.03 | 6.20 | 30.91 | 61.40 | 93.62 | 129.23 | 169.19 | 225.22 | 249.61 |
| | 0.090 | 2.62 | 0.27 | 1.81 | 3.20 | 4.41 | 5.62 | 6.65 | 7.38 | 39.97 | 65.15 | 104.28 | 149.85 | 208.70 | 243.93 | 244.39 | 116.71 | 6.17 | 3.71 | 1.04 | 3.25 | 4.79 | 31.76 | 61.45 | 92.80 | 127.46 | 166.35 | 220.21 | 243.95 |
| | 0.135 | 4.36 | 1.79 | 0.47 | 1.36 | 2.54 | 3.73 | 4.75 | 8.04 | 40.43 | 65.15 | 103.57 | 148.32 | 206.11 | 240.23 | 240.69 | 113.43 | 7.23 | 4.81 | 1.65 | 1.41 | 5.88 | 32.37 | 61.51 | 92.30 | 126.33 | 164.53 | 216.94 | 240.25 |
| | 0.180 | 5.61 | 3.09 | 1.34 | 0.28 | 1.92 | 2.93 | 3.59 | 8.11 | 39.84 | 64.07 | 101.72 | 145.57 | 202.20 | 235.56 | 236.00 | 110.94 | 7.43 | 4.99 | 2.31 | 0.82 | 5.99 | 31.94 | 60.51 | 90.68 | 124.03 | 161.46 | 212.73 | 235.58 |
| | 0.248 | 6.76 | 4.25 | 2.50 | 1.92 | 0.24 | 1.16 | 2.16 | 10.02 | 41.75 | 65.98 | 103.62 | 147.47 | 204.09 | 236.67 | 237.12 | 109.00 | 9.23 | 6.86 | 3.78 | 1.18 | 7.90 | 33.85 | 62.41 | 92.58 | 125.93 | 163.35 | 213.85 | 236.69 |
| | 0.315 | 7.86 | 5.36 | 3.63 | 2.91 | 1.17 | 0.39 | 0.98 | 10.94 | 42.37 | 66.37 | 103.67 | 147.12 | 203.22 | 235.09 | 235.54 | 106.99 | 10.15 | 7.80 | 4.79 | 2.28 | 8.84 | 34.55 | 62.84 | 92.73 | 125.77 | 162.85 | 212.49 | 235.12 |
| | 0.428 | 8.75 | 6.28 | 4.57 | 3.53 | 2.12 | 0.98 | 0.30 | 11.45 | 42.57 | 66.33 | 103.25 | 146.25 | 201.79 | 233.10 | 233.54 | 105.30 | 10.73 | 8.34 | 5.58 | 3.21 | 9.37 | 34.82 | 62.83 | 92.42 | 125.13 | 161.83 | 210.72 | 233.12 |
| | 0.585 | 7.96 | 5.97 | 6.92 | 7.12 | 8.80 | 9.69 | 10.25 | 0.46 | 27.87 | 49.14 | 82.21 | 120.72 | 170.45 | 202.59 | 202.98 | 104.54 | 9.47 | 7.30 | 5.63 | 7.84 | 2.86 | 20.93 | 46.02 | 72.51 | 101.80 | 134.67 | 182.55 | 202.61 |
| | 0.765 | 25.53 | 26.66 | 27.45 | 27.61 | 28.94 | 29.64 | 30.08 | 21.99 | 0.20 | 16.79 | 42.88 | 73.27 | 112.52 | 146.68 | 146.99 | 104.49 | 22.54 | 24.19 | 26.33 | 28.18 | 23.46 | 5.47 | 14.32 | 35.23 | 58.34 | 84.28 | 130.86 | 146.69 |
| | 0.990 | 45.21 | 46.41 | 47.27 | 47.43 | 48.85 | 49.60 | 50.07 | 41.43 | 17.93 | 0.38 | 27.99 | 61.61 | 103.43 | 146.13 | 146.50 | 129.57 | 42.02 | 43.77 | 46.07 | 48.04 | 43.00 | 23.78 | 4.47 | 19.70 | 44.40 | 74.24 | 130.07 | 146.17 |
| | 1.395 | 89.81 | 91.29 | 92.34 | 92.54 | 94.29 | 95.21 | 95.79 | 85.17 | 56.29 | 34.38 | 0.43 | 41.45 | 92.84 | 159.02 | 159.48 | 193.48 | 85.89 | 88.05 | 90.87 | 93.29 | 87.10 | 63.48 | 39.76 | 11.93 | 20.30 | 56.98 | 139.28 | 159.07 |
| | 1.890 | 183.25 | 185.34 | 186.82 | 187.11 | 189.58 | 190.88 | 191.69 | 176.69 | 135.90 | 106.94 | 58.54 | 1.17 | 72.82 | 190.52 | 191.10 | 329.71 | 177.71 | 180.76 | 184.74 | 188.16 | 179.42 | 146.05 | 114.71 | 75.39 | 31.04 | 21.96 | 161.30 | 190.55 |
| | 2.500 | 494.70 | 498.73 | 501.59 | 502.14 | 506.91 | 509.43 | 511.00 | 482.01 | 403.19 | 346.83 | 253.30 | 140.58 | 4.13 | 283.79 | 284.90 | 777.70 | 483.99 | 489.88 | 497.57 | 504.19 | 487.28 | 422.81 | 361.85 | 285.87 | 200.16 | 101.12 | 227.57 | 283.84 |
| | 3.289 | 93.87 | 93.87 | 93.87 | 93.87 | 93.88 | 93.88 | 93.88 | 93.88 | 93.88 | 93.85 | 93.79 | 93.72 | 93.63 | 93.60 | 93.60 | 93.88 | 93.86 | 93.87 | 93.87 | 93.87 | 93.87 | 93.87 | 93.86 | 93.81 | 93.75 | 93.69 | 93.63 | 93.60 |
| | 4.386 | 94.22 | 94.23 | 94.24 | 94.25 | 94.25 | 94.25 | 94.26 | 94.26 | 94.26 | 94.18 | 94.04 | 93.87 | 93.66 | 93.60 | 93.59 | 94.27 | 94.22 | 94.23 | 94.24 | 94.25 | 94.25 | 94.25 | 94.20 | 94.09 | 93.96 | 93.81 | 93.68 | 93.60 |
| 2x Gain | 0.068 | 93.60 | 93.60 | 93.60 | 93.60 | 93.60 | 93.60 | 93.60 | 93.60 | 93.60 | 93.60 | 93.60 | 93.60 | 93.60 | 93.60 | 93.60 | 93.60 | 93.60 | 93.60 | 93.60 | 93.60 | 93.60 | 93.60 | 93.60 | 93.60 | 93.60 | 93.60 | 93.60 | 93.60 |
| | 0.090 | 4.53 | 6.24 | 7.45 | 7.85 | 9.70 | 10.77 | 11.47 | 11.34 | 34.21 | 59.69 | 99.28 | 145.40 | 204.96 | 243.10 | 243.57 | 124.35 | 0.67 | 2.64 | 5.79 | 8.55 | 7.92 | 25.90 | 55.94 | 87.67 | 122.74 | 162.10 | 219.10 | 243.13 |
| | 0.135 | 2.00 | 3.68 | 4.87 | 5.22 | 7.08 | 8.13 | 8.79 | 8.57 | 36.04 | 61.05 | 99.92 | 145.20 | 203.67 | 240.15 | 240.61 | 119.64 | 2.63 | 0.40 | 3.21 | 5.95 | 5.22 | 27.88 | 57.37 | 88.53 | 122.96 | 161.60 | 216.59 | 240.17 |
| | 0.180 | 3.40 | 0.88 | 1.65 | 2.31 | 3.83 | 4.87 | 5.70 | 6.60 | 38.76 | 63.46 | 101.86 | 146.58 | 204.34 | 239.10 | 239.56 | 115.02 | 5.60 | 3.16 | 0.49 | 2.71 | 4.23 | 30.70 | 59.83 | 90.60 | 124.61 | 162.78 | 215.82 | 239.12 |
| | 0.248 | 5.70 | 3.16 | 1.40 | 0.83 | 1.18 | 2.29 | 3.30 | 8.99 | 40.97 | 65.37 | 103.31 | 147.49 | 204.54 | 237.81 | 238.26 | 110.94 | 8.19 | 5.81 | 2.70 | 0.31 | 6.86 | 33.01 | 61.78 | 92.18 | 125.78 | 163.49 | 214.82 | 237.83 |
| | 0.315 | 5.47 | 4.33 | 5.42 | 5.63 | 7.43 | 8.39 | 8.98 | 3.25 | 31.84 | 54.63 | 90.04 | 131.28 | 184.54 | 218.16 | 218.58 | 109.96 | 7.09 | 4.76 | 3.89 | 6.40 | 0.25 | 24.41 | 51.28 | 79.65 | 111.02 | 146.22 | 196.69 | 218.18 |
| | 0.428 | 21.85 | 23.08 | 23.94 | 24.11 | 25.56 | 26.33 | 26.81 | 17.99 | 5.97 | 24.26 | 52.69 | 85.80 | 128.56 | 163.39 | 163.73 | 107.88 | 18.59 | 20.38 | 22.72 | 24.73 | 19.60 | 0.51 | 21.57 | 44.35 | 69.53 | 97.79 | 146.16 | 163.41 |
| | 0.585 | 40.61 | 41.76 | 42.57 | 42.73 | 44.09 | 44.80 | 45.25 | 37.01 | 14.59 | 4.27 | 30.87 | 63.04 | 102.93 | 142.21 | 142.63 | 121.09 | 37.57 | 39.24 | 41.43 | 43.31 | 38.51 | 20.17 | 0.27 | 21.61 | 45.98 | 75.09 | 126.99 | 142.34 |
| | 0.765 | 72.61 | 73.96 | 74.91 | 75.09 | 76.68 | 77.52 | 78.05 | 68.38 | 42.10 | 22.04 | 10.87 | 48.59 | 95.37 | 151.49 | 151.99 | 166.98 | 69.04 | 71.00 | 73.57 | 75.78 | 70.14 | 48.64 | 25.33 | 0.72 | 28.58 | 62.73 | 133.65 | 151.64 |
| | 0.990 | 125.35 | 127.03 | 128.22 | 128.45 | 130.44 | 131.49 | 132.15 | 120.06 | 87.19 | 62.10 | 23.10 | 25.04 | 83.54 | 167.78 | 168.40 | 243.36 | 120.88 | 123.34 | 126.54 | 129.30 | 122.25 | 95.37 | 67.41 | 35.73 | 1.04 | 42.72 | 145.46 | 167.97 |
| | 1.395 | 242.93 | 245.42 | 247.19 | 247.53 | 250.47 | 252.03 | 253.00 | 235.10 | 186.44 | 153.71 | 95.98 | 26.13 | 62.50 | 212.65 | 213.33 | 417.63 | 236.32 | 239.96 | 244.71 | 248.79 | 238.36 | 198.56 | 162.99 | 116.08 | 63.18 | 1.51 | 177.65 | 212.68 |
| | 1.890 | 85.02 | 84.88 | 84.79 | 84.72 | 84.65 | 84.59 | 84.54 | 84.53 | 84.57 | 85.65 | 87.68 | 90.13 | 93.16 | 94.12 | 94.15 | 84.36 | 85.11 | 84.98 | 84.84 | 84.71 | 84.70 | 84.70 | 85.33 | 86.97 | 88.83 | 91.04 | 92.96 | 94.13 |
| | 4.386 | 94.05 | 94.06 | 94.07 | 94.07 | 94.07 | 94.08 | 94.08 | 94.08 | 94.08 | 94.02 | 93.92 | 93.80 | 93.64 | 93.60 | 93.60 | 94.09 | 94.05 | 94.06 | 94.06 | 94.07 | 94.07 | 94.07 | 94.04 | 93.96 | 93.86 | 93.75 | 93.66 | 93.60 |

**Figure 9 Object-based cross-calibration MAPE for the green band for various exposure and gain combinations, where the acceptable range is marked by the back box.**

Target images

| Reference image | Exposure time (ms) | 1x Gain | | | | | | | | | | | | | | | 2x Gain | | | | | | | | | | | | |
|---|---|---|---|---|---|---|---|---|---|---|---|---|---|---|---|---|---|---|---|---|---|---|---|---|---|---|---|---|---|
| | | 0.068 | 0.090 | 0.135 | 0.180 | 0.248 | 0.315 | 0.428 | 0.585 | 0.765 | 0.990 | 1.395 | 1.890 | 2.500 | 3.289 | 4.386 | 0.068 | 0.090 | 0.135 | 0.180 | 0.248 | 0.315 | 0.428 | 0.585 | 0.765 | 0.990 | 1.395 | 1.890 | 4.386 |
| 1x Gain | 0.068 | 1.32 | 0.75 | 1.06 | 2.16 | 3.54 | 3.73 | 3.75 | 3.40 | 3.29 | 3.27 | 10.07 | 101.81 | 117.11 | 159.55 | 128.87 | 3.83 | 2.35 | 1.58 | 2.18 | 2.70 | 2.63 | 2.26 | 15.86 | 46.58 | 101.06 | 138.75 | 105.61 | 69.91 |
| | 0.090 | 1.32 | 0.29 | 0.92 | 1.47 | 2.85 | 3.04 | 3.06 | 2.71 | 2.60 | 2.58 | 10.70 | 101.43 | 117.19 | 159.42 | 128.09 | 4.48 | 3.01 | 2.25 | 1.50 | 2.01 | 1.94 | 1.57 | 15.94 | 47.02 | 101.22 | 138.72 | 104.94 | 68.88 |
| | 0.135 | 1.45 | 0.40 | 0.94 | 1.50 | 2.76 | 3.07 | 3.09 | 2.74 | 2.62 | 2.61 | 11.02 | 101.82 | 117.87 | 160.24 | 128.45 | 4.79 | 3.32 | 2.55 | 1.56 | 2.04 | 1.97 | 1.51 | 16.23 | 47.46 | 101.85 | 139.47 | 105.22 | 68.81 |
| | 0.180 | 2.26 | 1.46 | 1.55 | 0.60 | 1.70 | 1.55 | 1.55 | 1.21 | 1.24 | 1.09 | 11.50 | 100.01 | 116.31 | 157.87 | 125.86 | 5.39 | 4.03 | 3.19 | 0.72 | 0.91 | 0.89 | 0.48 | 16.18 | 47.24 | 100.59 | 137.50 | 103.08 | 66.81 |
| | 0.248 | 3.74 | 2.72 | 2.74 | 1.69 | 0.17 | 0.90 | 1.03 | 0.83 | 0.64 | 1.38 | 13.15 | 99.96 | 117.55 | 158.95 | 125.04 | 7.06 | 5.62 | 4.87 | 2.26 | 1.91 | 1.38 | 1.38 | 17.79 | 48.75 | 101.89 | 138.65 | 102.34 | 64.86 |
| | 0.315 | 3.67 | 2.96 | 2.99 | 1.52 | 0.36 | 0.67 | 0.68 | 0.50 | 0.58 | 1.04 | 12.69 | 98.95 | 116.11 | 157.12 | 123.92 | 6.69 | 5.39 | 4.49 | 1.97 | 1.63 | 1.23 | 1.51 | 17.30 | 47.96 | 100.60 | 137.01 | 101.44 | 64.58 |
| | 0.428 | 3.65 | 2.96 | 2.99 | 1.53 | 0.88 | 0.88 | 0.53 | 0.49 | 0.81 | 0.55 | 12.06 | 98.26 | 114.88 | 155.64 | 123.31 | 6.32 | 5.04 | 3.91 | 1.74 | 1.22 | 1.08 | 1.52 | 16.82 | 47.13 | 99.46 | 135.66 | 100.96 | 64.76 |
| | 0.585 | 3.33 | 2.64 | 2.67 | 1.21 | 0.82 | 0.77 | 0.64 | 0.34 | 0.59 | 0.55 | 12.21 | 98.92 | 115.70 | 156.73 | 124.11 | 6.27 | 4.97 | 4.00 | 1.58 | 1.17 | 0.90 | 1.19 | 16.85 | 47.50 | 100.18 | 136.61 | 101.61 | 65.12 |
| | 0.765 | 3.40 | 2.54 | 2.57 | 1.27 | 0.49 | 0.75 | 0.83 | 0.52 | 0.45 | 0.90 | 12.63 | 99.54 | 116.68 | 157.95 | 124.72 | 6.56 | 5.17 | 4.37 | 1.78 | 1.44 | 0.95 | 1.14 | 17.26 | 48.11 | 101.08 | 137.72 | 102.10 | 65.13 |
| | 0.990 | 3.19 | 2.51 | 2.53 | 1.12 | 1.36 | 1.03 | 0.65 | 0.60 | 0.98 | 0.24 | 11.62 | 98.50 | 114.78 | 155.69 | 123.82 | 5.87 | 4.58 | 3.44 | 1.28 | 0.85 | 0.92 | 1.06 | 16.40 | 46.81 | 99.31 | 135.63 | 101.39 | 65.45 |
| | 1.395 | 8.74 | 9.35 | 9.60 | 10.21 | 11.72 | 11.42 | 10.92 | 10.98 | 11.29 | 10.49 | 0.33 | 86.78 | 93.08 | 129.99 | 113.82 | 5.46 | 6.71 | 7.38 | 9.70 | 10.02 | 10.49 | 10.54 | 11.76 | 31.75 | 79.12 | 111.90 | 93.59 | 69.55 |
| | 1.890 | 62.42 | 62.27 | 62.27 | 61.94 | 61.68 | 61.60 | 61.59 | 61.67 | 61.70 | 61.70 | 61.80 | 61.80 | 66.86 | 75.44 | 70.32 | 62.88 | 62.62 | 62.20 | 61.94 | 61.82 | 61.84 | 61.94 | 61.95 | 61.97 | 62.19 | 70.85 | 69.18 | 66.17 |
| | 2.500 | 55.16 | 55.51 | 55.66 | 56.02 | 56.89 | 56.72 | 56.43 | 56.46 | 56.64 | 56.18 | 50.08 | 46.96 | 15.42 | 32.96 | 83.68 | 53.24 | 53.98 | 54.37 | 55.72 | 55.90 | 56.18 | 56.21 | 47.68 | 31.64 | 15.20 | 22.56 | 71.58 | 90.49 |
| | 3.289 | 83.45 | 83.84 | 84.00 | 84.40 | 85.37 | 85.18 | 84.86 | 84.89 | 85.10 | 84.58 | 77.80 | 74.32 | 21.07 | 21.30 | 98.75 | 81.31 | 82.14 | 82.57 | 84.07 | 84.27 | 84.58 | 84.61 | 75.12 | 57.28 | 33.40 | 17.57 | 85.29 | 122.75 |
| | 4.386 | 77.30 | 77.23 | 77.23 | 77.08 | 76.96 | 76.92 | 76.92 | 76.95 | 76.96 | 76.97 | 77.01 | 77.01 | 79.35 | 83.31 | 80.94 | 77.51 | 77.39 | 77.20 | 77.08 | 77.02 | 77.03 | 77.08 | 77.08 | 77.09 | 77.19 | 81.19 | 80.42 | 79.03 |
| 2x Gain | 0.068 | 3.78 | 4.48 | 4.77 | 5.47 | 7.21 | 6.86 | 6.29 | 6.35 | 6.72 | 5.79 | 6.25 | 100.91 | 113.15 | 155.54 | 129.47 | 0.91 | 1.76 | 3.05 | 4.99 | 5.29 | 5.79 | 5.86 | 14.76 | 42.71 | 97.12 | 134.76 | 106.23 | 73.62 |
| | 0.090 | 2.37 | 2.98 | 3.27 | 3.97 | 5.69 | 5.35 | 4.85 | 4.84 | 5.20 | 4.30 | 7.68 | 100.56 | 113.90 | 156.01 | 128.34 | 1.82 | 1.28 | 1.83 | 3.56 | 3.89 | 4.29 | 4.35 | 14.96 | 43.91 | 97.97 | 135.37 | 105.26 | 71.68 |
| | 0.135 | 2.18 | 2.19 | 2.47 | 3.15 | 4.83 | 4.49 | 3.94 | 4.00 | 4.35 | 3.46 | 8.22 | 98.28 | 111.87 | 152.97 | 125.10 | 3.03 | 1.85 | 0.23 | 2.58 | 2.93 | 3.46 | 3.52 | 14.74 | 43.57 | 96.33 | 132.82 | 102.57 | 69.23 |
| | 0.180 | 2.18 | 1.48 | 1.68 | 0.85 | 2.24 | 1.93 | 1.58 | 1.42 | 1.77 | 1.06 | 10.87 | 99.28 | 115.03 | 156.33 | 125.19 | 4.99 | 3.69 | 2.61 | 0.77 | 0.88 | 1.14 | 0.93 | 15.74 | 46.39 | 99.41 | 136.08 | 102.55 | 66.96 |
| | 0.248 | 2.67 | 1.97 | 2.00 | 0.88 | 1.89 | 1.55 | 1.14 | 1.06 | 1.41 | 0.61 | 11.16 | 98.82 | 114.77 | 155.85 | 124.46 | 5.37 | 4.07 | 2.94 | 0.77 | 0.62 | 0.95 | 0.76 | 15.95 | 46.49 | 99.23 | 135.71 | 101.94 | 66.26 |
| | 0.315 | 2.63 | 1.91 | 1.94 | 0.64 | 1.37 | 1.16 | 1.08 | 0.75 | 0.90 | 0.62 | 11.76 | 99.58 | 116.04 | 157.39 | 125.17 | 5.69 | 4.39 | 3.49 | 0.93 | 0.81 | 0.78 | 0.48 | 16.40 | 47.32 | 100.40 | 137.12 | 102.50 | 66.17 |
| | 0.428 | 2.50 | 1.51 | 1.53 | 0.72 | 1.41 | 1.53 | 1.54 | 1.20 | 1.09 | 1.08 | 11.92 | 100.44 | 117.10 | 158.80 | 126.23 | 5.78 | 4.33 | 3.58 | 0.96 | 0.95 | 0.90 | 0.36 | 16.60 | 47.79 | 101.32 | 138.36 | 103.37 | 66.67 |
| | 0.585 | 9.61 | 10.17 | 10.41 | 10.97 | 12.36 | 12.09 | 11.63 | 11.68 | 11.97 | 11.23 | 3.46 | 79.85 | 84.44 | 118.54 | 105.45 | 7.45 | 7.89 | 8.36 | 10.50 | 10.79 | 11.23 | 11.28 | 10.56 | 27.78 | 71.55 | 101.82 | 86.76 | 65.79 |
| | 0.765 | 27.16 | 27.60 | 27.78 | 28.23 | 29.33 | 29.11 | 28.75 | 28.79 | 29.02 | 28.43 | 20.79 | 59.06 | 47.02 | 73.91 | 87.08 | 24.75 | 25.68 | 26.17 | 27.86 | 28.09 | 28.43 | 28.47 | 17.77 | 10.89 | 36.85 | 60.73 | 72.34 | 71.46 |
| | 0.990 | 40.69 | 41.01 | 41.14 | 41.45 | 42.23 | 42.07 | 41.82 | 41.85 | 42.01 | 41.59 | 36.20 | 39.83 | 19.77 | 38.48 | 73.63 | 38.99 | 39.65 | 40.00 | 41.19 | 41.35 | 41.59 | 41.62 | 34.07 | 19.89 | 15.06 | 29.28 | 62.92 | 71.95 |
| | 1.395 | 67.13 | 67.50 | 67.65 | 68.02 | 68.92 | 68.74 | 68.44 | 68.48 | 68.67 | 68.18 | 61.89 | 58.67 | 13.88 | 24.84 | 89.44 | 65.15 | 65.92 | 66.32 | 67.71 | 67.90 | 68.18 | 68.22 | 59.41 | 42.86 | 20.90 | 18.36 | 76.94 | 103.60 |
| | 1.890 | 71.54 | 71.42 | 71.43 | 71.17 | 70.97 | 70.91 | 70.90 | 70.96 | 70.98 | 70.98 | 71.06 | 71.06 | 74.97 | 81.60 | 77.64 | 71.90 | 71.69 | 71.38 | 71.17 | 71.08 | 71.09 | 71.17 | 71.18 | 71.19 | 71.36 | 78.06 | 76.77 | 74.44 |
| | 4.386 | 84.21 | 84.22 | 84.22 | 84.23 | 84.24 | 84.24 | 84.24 | 84.24 | 84.24 | 84.24 | 84.23 | 84.23 | 84.06 | 83.76 | 83.94 | 84.20 | 84.21 | 84.22 | 84.23 | 84.23 | 84.23 | 84.23 | 84.23 | 84.23 | 84.22 | 83.92 | 83.98 | 84.08 |

**Figure 10 Object-based cross-calibration MAPE for the red band for various exposure and gain combinations, where the acceptable range is marked by the back box.**

### 3.4 Reflectance Accuracy in Auto-Exposure and Fixed Exposure Orthomosaics

Figure 11 and Figure 12 compare ground-truth spectral reflectance with estimated reflectance of the in-field calibration targets after post-processing calibration of the 17 June 2021 and 16 June 2022 orthomosaics, respectively. The graphs contain the object-based ELM equation, $R^2$ based on linear regression, and MAPE after the ELM calibration. It should be noted that the white target reflectance in 2021 was only about 0.50, while in 2020 it was close to 0.85, thereby providing a much wider range to assess reflectance characteristics, especially in the red edge and NIR bands. In 2021, fixed exposure settings ensured higher spatial uniformity ($R^2$ = 0.99) and reduction in error by at least half that of auto-exposure. The NIR band was an exception, wherein the $R^2$ and MAPE values were similar for fixed and autoexposure orthomosaics. In 2022, the overall performance of the fixed exposure images was better than that of autoexposure, however not greatly different in the green and red bands. The red edge and NIR bands in 2022 had noticeably better $R^2$ and MAPE values for fixed exposure relative to autoexposure. In both datasets, the fixed exposure images' blue band (2021: $R^2$ = 0.99 and MAPE = 3.11; 2022: $R^2$ = 0.97 and MAPE = 5.91) clearly outperformed the autoexposure images (2021: $R^2$ = 0.75 and MAPE = 24.25; 2022: $R^2$ = 0.79 and MAPE = 25.06). The 2022 fixed exposure images did not achieve the same level or accuracy and spatial uniformity (precision) as the 2021 fixed exposure images in the blue, green, and red bands, but they performed better than autoexposure in both years.

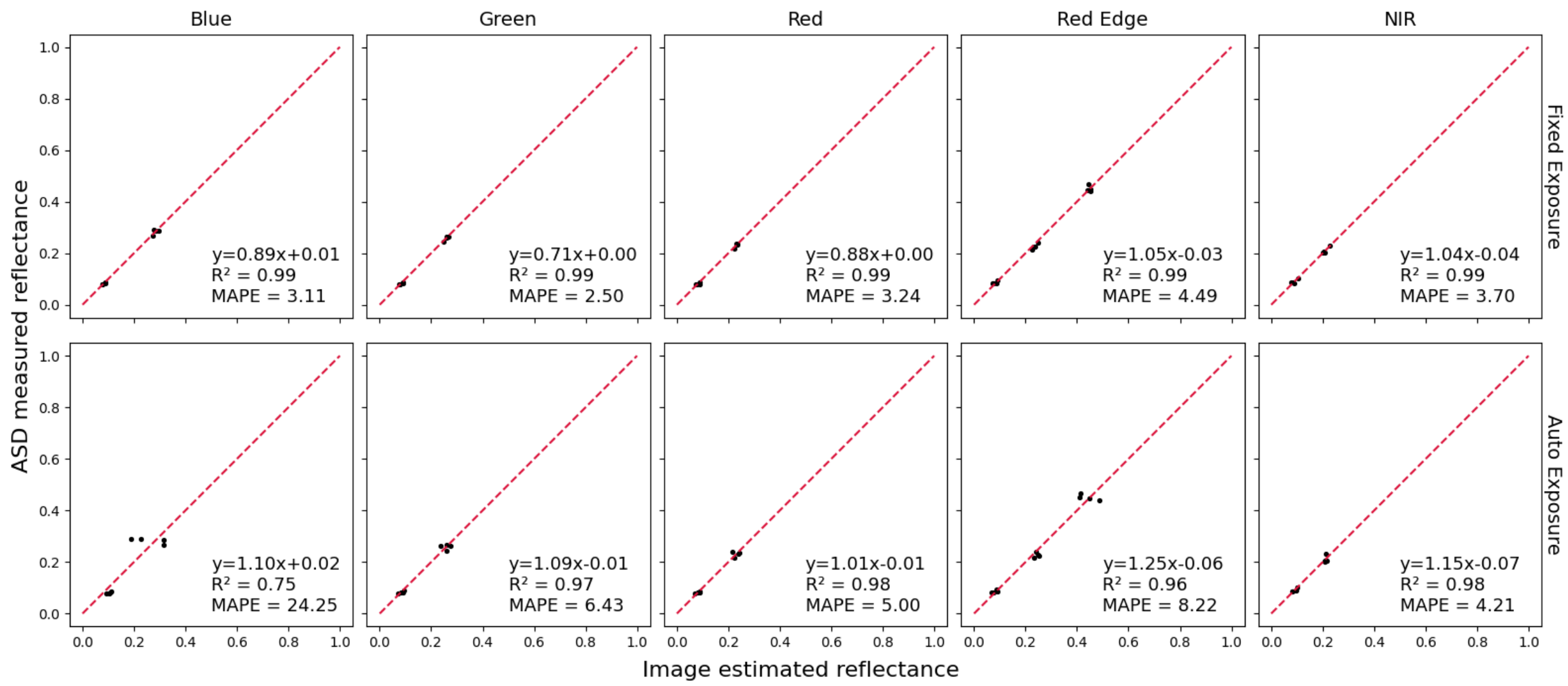

**Figure 11 Object-based ELM performed on the blue, green, red, red edge, and NIR (left to right) bands of fixed exposure (top) and auto-exposure (bottom) datasets acquired on 17 June 2021. The ELM performance is denoted by the empirical line equation, coefficient of determination ($R^2$) and MAPE.**

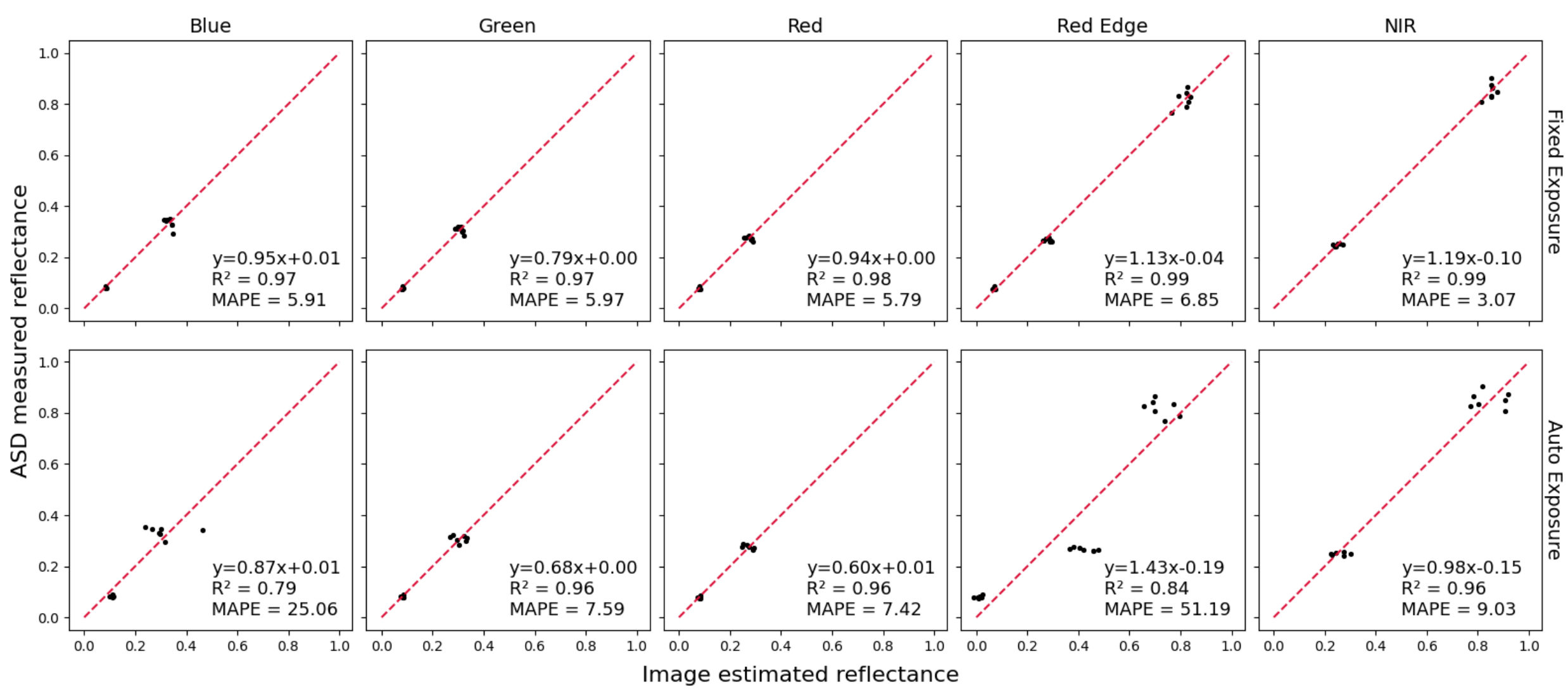

**Figure 12 Object-based ELM performed on the blue, green, red, red edge, and NIR (left to right) bands of fixed exposure (top) and auto-exposure (bottom) datasets acquired on 16 June 2022. The ELM performance is denoted by the empirical line equation, coefficient of determination ($R^2$) and MAPE.**

### 3.5 Vegetation Indices for Total N Estimation

Vegetation indices derived from 17 June 2021 and 16 June 2021 fixed and autoexposure orthomosiacs (Table 4) were used for total accumulated biomass N estimation. Figure 13 shows a comparison of the actual and predicted total N to compare the performance of VIs calculated from uncalibrated and calibrated orthomosaics of fixed and autoexposure datasets. Noticeably higher $R^2$, lower MAPE, and significant correlation ($p < 0.05$) with total N were observed for uncalibrated and calibrated fixed exposure VIs compared to autoexposure VIs. NDVI and RDVI, which had similar equations, did not differ significantly in terms of $R^2$ and MAPE between fixed exposure and uncalibrated autoexposure. Calibrated auto-exposure VIs, which had the poorest correlation with total N among the four, showed no significant correlation with total N except for $CI_{green}$ and RDVI. It was also observed that calibrated autoexposure $CI_{rededge}$ and NDRE, based on the red edge and NIR bands, respectively, had the lowest $R^2$ values. Although calibrated fixed exposure had the best $R^2$ values among all VIs, the improvement in MAPE compared to calibrated and uncalibrated autoexposure was only 2 to 6 %).

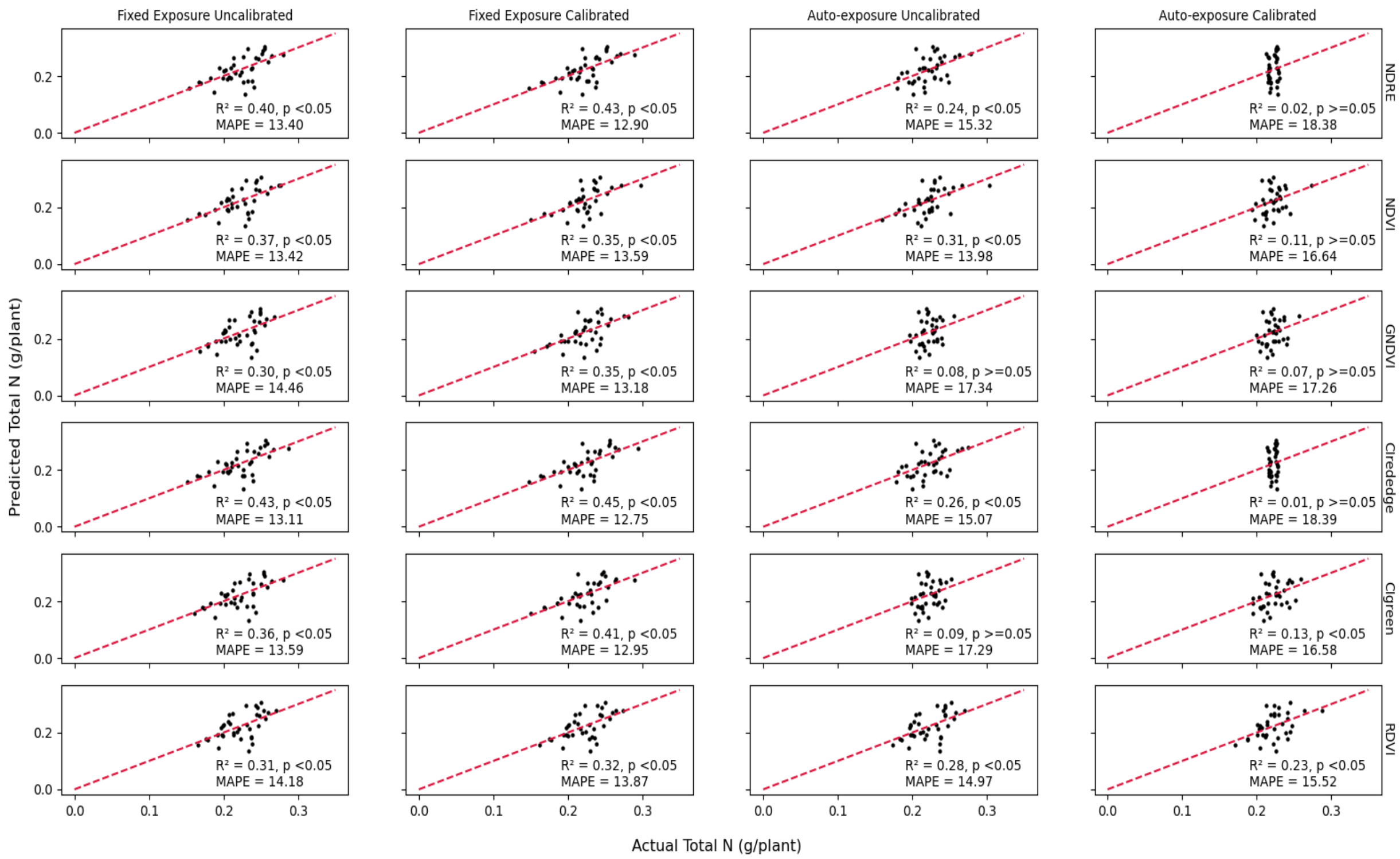

**Figure 13 Comparison of actual total N content in the nitrogen management plots with estimated total N from six VIs (rows top to bottom: NDRE, NDVI, GNDVI, $CI_{rededge}$, $CI_{green}$, and RDVI) collected on 17 June 2021 and 16 June 2022 using four different methods (columns left to right: fixed exposure, fixed exposure calibrated, auto-exposure, and auto-exposure calibrated).**

# 4. Discussion

## 4.1 Radiometric calibration and object-based ELM

Radiometric calibration is a necessary step in image-based remote sensing to standardize and improve the accuracy of data collected from different sensors, at different times, and different locations. Many studies have used reference panel-based and panel-less calibration of UAV-collected aerial multispectral images, but panel-based multi-point ELM has proven to be the most reliable calibration method [18]. The CRP-based 1-point ELM calibration, which is widely recommended by camera manufacturers, is prone to calibration inaccuracies because it assumes the calibration equation regression line passes through the origin, but that assumption may not be valid due to the influence of lighting and atmospheric conditions that may result in a non-zero intercept. Also, the wear and tear on a camera's lens (distortion parameters) and electronic components can affect its sensitivity over time. The various multi-point ELM methods listed in the Introduction section used two or more reflectance levels to accurately determine the slope and intercept for UAV image calibration. In this study, we proposed an object-based ELM calibration, a type of multi-point ELM, in which the calibration range was selected to focus on crop canopy and soil reflectances. This meant the blue, green, and red bands were calibrated within a narrow region bounded by the low (~0.07) and medium (~0.30) reflectance targets, as canopy and soil reflectance in these bands did not exceed 0.30. The NIR band was calibrated between medium and high (~0.80) reflectance, and the red edge band was calibrated with all targets to cover the wider range of expected soil and canopy reflectance in these bands. The object-based ELM complemented the fixed exposure-settings used in this study that optimized on the reflectance range of soil and canopy objects (Table 5). Therefore, if the exposure setting were chosen to capture the entire reflectance range from 0.00 to 1.00, object-based ELM in that case will be the same as multi-point ELM, where all the targets (B, G, and W) are included for calibration. Object-based ELM worked reasonably well on the auto-exposure data set, too, because the auto-exposure settings optimized the dynamic range of the images based on the soil and canopy objects on ground. Here again, the object-based ELM complemented the exposure settings used to acquire images.

### 4.2 Problems with auto-exposure

Most studies on radiometric calibration of UAV-borne multispectral cameras for precision agriculture used the default auto-exposure settings, and calibration performance was assessed based on errors in estimating reflectance of known reference targets. While autoexposure settings help to minimize loss of low-intensity data through histogram equalization, the influence of exposure settings on radiance or reflectance estimation accuracy, especially in scenes with drastic variations in radiance intensities, remained heretofore unexplored. Comparing the estimated and actual reflectances of calibration targets in the orthomosaic showed reasonable accuracy (MAPE < 10 %) and spatial uniformity ($R^2 > 0.96$) for autoexposure in the green, red, and NIR bands, and relatively high MAPE in the blue and red edge bands (Figure 11:bottom rows and Figure 12:bottom rows). However, due to significant variations in exposure time and gain when capturing different objects like canopy, soil, and in-field reflectance calibration targets (Figure 2 and Figure 3), we probed the effectiveness of the equations recommended by the camera manufacturer (equations 1 to 7) in providing consistent measures on radiance or reflectance under auto-exposure settings. The first observation was that the blue and red edge bands, which had lower $R^2$ and higher MAPE, were also prone to relatively higher variations in exposure time and gain when autoexposure was used. Further, the band-wise cross-calibration MAPE matrix (Figure 6 to Figure 10) quantified the radiometric MAPE from applying object-based ELM derived from one exposure setting to images with different exposure settings. Here, it was observed that the cross-calibration MAPE remained low (< 5 % or <10 % for red edge) only when the exposure values were within the ideal exposure upper limits (Table 5). In actuality, however, the auto-exposure settings of the camera were commonly beyond the prescribed upper limits when capturing the different objects (Figure 2 and Figure 3). It was observed that higher exposure times progressed towards saturation of target reflectance values at 0.50 ratio (Figure 4 and Figure 5), resulting in unrecoverable loss of details. Therefore, the non-linear relationship between estimated and actual reflectance in low reflecting bands in other studies [20,21] can be attributed to reflectance of the high reflecting targets tending towards saturation in response to the higher exposure times. These results and observations highlight the importance of exposure time on radiometric accuracy, which can be compromised if exposure settings (1) change drastically from scene-to-scene and object-to-object and (2) are beyond the ideal limits. Since it is difficult to control exposure variations in autoexposure flights, we recommend the usage of fixed exposure settings to have greater control over the radiometric accuracy of the images.

### 4.3 Fixed exposure for UAV image acquisition

Two sets of ideal exposure time ranges are listed in Table 5 for full-scale multipoint and object-based ELM, respectively. However, full-scale reflectance (0 – 1) calibration is not required under field conditions, especially in the visible spectral bands where canopy and soil reflectance generally do not exceed 0.30. The ideal exposure values for object-based ELM were determined such that the estimated reflectances of the preferred targets were as close as possible to ground-truth (straight line), while ensuring reflectances of other targets did not saturate at 0.50 ratio. The RedEdge-3 exposure settings during UAV missions were set close to the upper limits of the ideal range at 1x gain to capture low reflecting objects with sufficient radiometric resolution while also avoiding saturation. The actual exposure times for the green and NIR bands in 2021 and blue and green bands in 2022 (Table 2) were marginally more (one setting higher) than the prescribed ideal exposure upper limits (Table 5). Since the ideal exposure settings were determined at 30 m AGL under clear sky conditions, and they may vary for different altitudes and illumination conditions.

Since soil and canopy were the objects of interest, fixed exposure ensured that radiometric readings were not distorted by the presence of other (highly reflective) objects within the field. Fixed exposure ensured that the raw DNs, radiance, and reflectance estimated for similar objects stayed consistent throughout the image acquisition process. Fixed exposure along with object-based ELM produced better spatial uniformity (higher $R^2$) and accuracy (lower MAPE) in target reflectance estimation for all bands, and it also produced VIs with better correlation to total N (g/plant) than did auto-exposure (Figure 13). The results here demonstrate that fixed exposure time and gain settings improved radiometric accuracy, which is important for quantitative remote sensing.

4.4 Future work

Ihe reflectance saturation occurred at 0.50 instead of the maximum reflectance value (1.00) because of the equations that converted DNs to radiance or reflectance (equations 1 to 7) and the subsequent CRP-based calibration. This study did not explore the mathematical reason behind reflectance saturation at half the maximum reflectance, so exploration of that phenomenon is left to future study. It should be noted that although the VIs from calibrated fixed exposure images had higher correlation with total N than VIs from auto-exposure images (uncalibrated and calibrated), the R2 values were still less than 0.5 in most cases. This was because the image datasets for both years were acquired early in the cotton-growth cycle when the canopy had not attained full closure, and plants had thus just started responding to the different N treatments. In future, these experiments should be repeated to determine the effect of exposure settings on radiometric calibration accuracy in mid growth stages. Also, this study was conducted with three calibration targets to select the calibration range and ideal exposure time. It would be interesting to include more color graded calibration targets in between the existing black, gray, and white targets, which would increase the resolution of the calibration panels to precisely fine-tune the calibration range and exposure time at different growth stages for different types of plants. We carried out the fixed exposure experiments with 1x gain only, and the ideal exposure times listed in Table 5 are relevant only for the RedEdge-3 camera used in this study. Future work should investigate in detail the effect of gain settings on radiometric calibration. It would also be interesting to develop ways to automatically choose exposure time and gain settings for each band based on conditions prevalent in the field.

## 5. Conclusions

This study was conducted to analyze the impact of exposure time on radiometric calibration accuracy of UAV-based multispectral images, specifically those obtained from the Micasense RedEdge-3 camera. It was observed that exposure time had a significant impact on the radiometric resolution and reflectance accuracy of image based reflectance estimates. The results also showed that autoexposure led to drastic variability in camera exposure settings, which can be detrimental to the spatiotemporal consistency in radiometric data. Hence, we concluded that the conventional way of using autoexposure settings for image acquisition is more prone to errors. The use of fixed exposure settings is recommended for canopy and soil reflectance estimation to avoid radiometric inaccuracies arising from drastic scene-to-scene or object-to-object variations in camera exposure time and gain.

## References

[1] Wang T, Thomasson JA, Yang C, Isakeit T, Nichols RL. Automatic classification of cotton root rot disease based on UAV remote sensing. Remote Sensing 2020;12(8):1310.

[2] Yeom J, Jung J, Chang A, Maeda M, Landivar J. Automated open cotton boll detection for yield estimation using unmanned aircraft vehicle (UAV) data. Remote Sensing 2018;10(12):1895.

[3] Osco LP, De Arruda, Mauro dos Santos, Junior JM, Da Silva NB, Ramos APM, Moryia ÉAS, et al. A convolutional neural network approach for counting and geolocating citrus-trees in UAV multispectral imagery. ISPRS Journal of Photogrammetry and Remote Sensing 2020;160:97-106.

[4] Yadav PK, Thomasson JA, Hardin R, Searcy SW, Braga-Neto U, Popescu SC, et al. Detecting volunteer cotton plants in a corn field with deep learning on UAV remote-sensing imagery. Comput.Electron.Agric. 2023;204:107551.

[5] Zhang L, Zhang H, Niu Y, Han W. Mapping maize water stress based on UAV multispectral remote sensing. Remote Sensing 2019;11(6):605.

[6] Bian J, Zhang Z, Chen J, Chen H, Cui C, Li X, et al. Simplified evaluation of cotton water stress using high resolution unmanned aerial vehicle thermal imagery. Remote Sensing 2019;11(3):267.

[7] Early estimation of nitrogen stress and uptake in cotton based on spectral and morphological features extracted from UAV multispectral images. Autonomous Air and Ground Sensing Systems for Agricultural Optimization and Phenotyping VII: SPIE; 2022.

[8] Zhou X, Kono Y, Win A, Matsui T, Tanaka TS. Predicting within-field variability in grain yield and protein content of winter wheat using UAV-based multispectral imagery and machine learning approaches. Plant Production Science 2021;24(2):137-51.

[9] Siegfried J, Adams CB, Rajan N, Hague S, Schnell R, Hardin R. Combining a cotton 'Boll Area Index'with in-season unmanned aerial multispectral and thermal imagery for yield estimation. Field Crops Res. 2023;291:108765.

[10] Xie C, Yang C. A review on plant high-throughput phenotyping traits using UAV-based sensors. Comput.Electron.Agric. 2020;178:105731.

[11] Shi Y, Thomasson JA, Murray SC, Pugh NA, Rooney WL, Shafian S, et al. Unmanned aerial vehicles for high-throughput phenotyping and agronomic research. PloS one 2016;11(7):e0159781.

[12] Frontera F, Smith MJ, Marsh S. Preliminary investigation into the geometric calibration of the micasense rededge-m multispectral camera. ISPRS-International Archives of the Photogrammetry, Remote Sensing and Spatial Information Sciences 2020;43.

[13] Mamaghani B, Salvaggio C. Multispectral sensor calibration and characterization for sUAS remote sensing. Sensors 2019;19(20):4453.

[14] Barker JB, Woldt WE, Wardlow BD, Neale CM, Maguire MS, Leavitt BC, et al. Calibration of a common shortwave multispectral camera system for quantitative agricultural applications. Precision Agriculture 2020;21:922-35.

[15] Cao H, Gu X, Wei X, Yu T, Zhang H. Lookup table approach for radiometric calibration of miniaturized multispectral camera mounted on an unmanned aerial vehicle. Remote Sensing 2020;12(24):4012.

[16] Mamaghani B, Salvaggio C. Comparative study of panel and panelless-based reflectance conversion techniques for agricultural remote sensing. arXiv preprint arXiv:1910.03734 2019.

[17] Guo Y, Senthilnath J, Wu W, Zhang X, Zeng Z, Huang H. Radiometric calibration for multispectral camera of different imaging conditions mounted on a UAV platform. Sustainability 2019;11(4):978.

[18] Poncet AM, Knappenberger T, Brodbeck C, Fogle Jr M, Shaw JN, Ortiz BV. Multispectral UAS data accuracy for different radiometric calibration methods. Remote Sensing 2019;11(16):1917.

[19] Comparison of reflectance calibration workflows for a uav-mounted multi-camera array system. 2021 IEEE International Geoscience and Remote Sensing Symposium IGARSS: IEEE; 2021.

[20] Deng L, Hao X, Mao Z, Yan Y, Sun J, Zhang A. A subband radiometric calibration method for UAV-based multispectral remote sensing. IEEE Journal of Selected Topics in Applied Earth Observations and Remote Sensing 2018;11(8):2869-80.

[21] Luo S, Jiang X, Yang K, Li Y, Fang S. Multispectral remote sensing for accurate acquisition of rice phenotypes: Impacts of radiometric calibration and unmanned aerial vehicle flying altitudes. Frontiers in Plant Science 2022;13.

[22] Nidamanuri RR, Zbell B. Existence of characteristic spectral signatures for agricultural crops–potential for automated crop mapping by hyperspectral imaging. Geocarto Int. 2012;27(2):103-18.

[23] Zhang H, Lan Y, Suh CP, Westbrook JK, Lacey R, Hoffmann WC. Differentiation of cotton from other crops at different growth stages using spectral properties and discriminant analysis. Transactions of the ASABE 2012;55(4):1623-30.

[24] Arafat SM, Aboelghar MA, Ahmed EF. Crop discrimination using field hyper spectral remotely sensed data. 2013.

[25] Bunnik N. Spectral reflectance characteristics of agricultural crops and application to crop growth monitoring. Advances in Space Research 1981;1(10):21-40.

[26] Olsson P, Vivekar A, Adler K, Garcia Millan VE, Koc A, Alamrani M, et al. Radiometric correction of multispectral uas images: Evaluating the accuracy of the parrot sequoia camera and sunshine sensor. Remote Sensing 2021;13(4):577.

[27] Bagnall GC, Thomasson JA, Yang C, Wang T, Han X, Sima C, et al. Uncrewed aerial vehicle radiometric calibration: A comparison of autoexposure and fixed-exposure images. The Plant Phenome Journal 2023;6(1):e20082.